\documentclass{article}

\usepackage{times}
\usepackage{graphicx} 
\usepackage{epsfig} 
\usepackage{subfigure} 

\usepackage{natbib}

\usepackage{algorithm}
\usepackage{algorithmic}

\usepackage{amsmath}
\usepackage{amssymb}
\usepackage{amsthm}
\usepackage{bm}
\usepackage{caption} 
\usepackage{epstopdf}
\usepackage[title]{appendix}

\usepackage{thmtools}

\newtheorem{theorem}{Theorem}
\newtheorem{remark}{Remark}

\declaretheoremstyle[%
  spaceabove=-6pt,%
  spacebelow=6pt,%
  headfont=\normalfont\itshape,%
  postheadspace=1em,%
  qed=\qedsymbol%
]{mystyle} 
\declaretheorem[name={Proof},style=mystyle,unnumbered,
]{prf}

\usepackage{hyperref}

\newcommand{\mbf}[1]{\mathbf{#1}}
\newcommand{\mbs}[1]{\boldsymbol{#1}}
\newcommand{\what}[1]{\widehat{#1}}
\newcommand{\wtilde}[1]{\widetilde{#1}}

\newcommand{\etal}{\textit{et al}. }
\newcommand{\x}{\mbf{x}}
\newcommand{\X}{\mbf{X}}
\newcommand{\y}{\mbf{y}}
\newcommand{\z}{\mbf{z}}
\newcommand{\indvar}{\mbf{u}}
\newcommand{\eps}{\varepsilon}
\newcommand{\epsvar}{\sigma^2_\varepsilon}
\newcommand{\pvec}{\mbs{\theta}}
\newcommand{\psub}{\theta}

\newcommand{\T}{\top}
\newcommand{\mvec}{\mbs{\mu}}
\newcommand{\covmat}{\mbs{\Sigma}}
\newcommand{\dataset}{\mathcal{D}}
\newcommand{\loss}{\ell}

\usepackage[accepted]{icml2018}

\icmltitlerunning{Composite Gaussian Processes: Scalable Computation and Performance Analysis}

\begin{document} 

\twocolumn[
\icmltitle{Composite Gaussian Processes: \\ Scalable Computation and Performance Analysis}



\icmlsetsymbol{equal}{*}

\begin{icmlauthorlist}
\icmlauthor{Xiuming Liu}{uu}
\icmlauthor{Dave Zachariah}{uu}
\icmlauthor{Edith C. H. Ngai}{uu}
\end{icmlauthorlist}

\icmlaffiliation{uu}{Uppsala University, Uppsala, Sweden}

\icmlcorrespondingauthor{Xiuming Liu and Dave Zachariah}{xiuming.liu@it.uu.se and dave.zachariah@it.uu.se}

\icmlkeywords{Scalable Gaussian Processes, Belief Updating, Recursive Bayesian Estimation, Information Theory}

\vskip 0.3in]



\printAffiliationsAndNotice{}  

\begin{abstract} 
Gaussian process (GP) models provide a powerful tool for prediction but are computationally prohibitive using large data sets. In such scenarios, one has to resort to approximate methods. We derive an approximation based on a composite likelihood approach using a general belief updating framework, which leads to a recursive computation of the predictor as well as of learning the hyper-parameters. We then provide an analysis of the derived composite GP model in predictive and information-theoretic terms. Finally, we evaluate the approximation with both synthetic data and a real-world application. 

\end{abstract}

\section{Introduction}\label{sec:introducion}

Regression is a fundamental problem in machine learning, signal processing, and system identification. In general, an input-output pair $(\x, y)$  can be described by
\begin{equation}
    y = f(\x) + \eps, 
\label{eq:model}
\end{equation}
where $f(\cdot)$ is an unknown regression function and $\eps$ is a zero-mean error term. Given a set of input-output pairs, the goal is to model $f(\cdot)$ and infer the value of $f(\x^*)$ at some test points.

Gaussian Processes (GPs) are a family of nonlinear and non-parametric models \cite{2006-Rasmussen-GPML}. Inferences based on GPs are conceptually straightforward and the model class provides an internal measure of uncertainties of the inferred quantities. Therefore GPs are widely applied in machine learning \cite{2006-Bishop-PatternRecognition}, time series analysis \cite{2011-Shumway-TimeSeries}, spatial statistics \cite{2015-Kroese-SpatialProcessSimulation}, and control systems \cite{2015-Deisenroth-GPforDataEfficientLearningAndControl}. 

A GP model of $f(\x)$ is specified by its mean and covariance functions, which are learned from data. After GP learning, the latent values of $f(\cdot)$ can be inferred from observed data. Both GP learning and inference rely on evaluating a likelihood function with available data. In case of a large set of data, this evaluation becomes computational prohibitive due to the requirement of inverting covariance matrices, which has a typical runtime on the order $\mathcal{O}(N^3)$, where $N$ is the size of training data. Thus the direct use of GP has been restricted to moderately sized data sets.

Considerable research efforts have investigated scalable GP models or approximations which can be divided into four broad categories: subset of data, low rank approximation of covariance matrices \cite{2000-Williams-NystromSpeedUpKernel}, sparse GPs \cite{2005-Quinonero-Candela-AUnifyingViewOfSparseGP, 2013-Hensman-GPBigData}, and product of experts (or Bayesian committee machines) \cite{2000-Tresp-BayesianCommitte,2015-deisenroth-distributedGP}. Many methods rely on fast heuristic searches \cite{2003-Seeger-FastForwardSelection,2009-Titsias-VariationalLearning} for an optimal subset of data or dimensions to, for instance, maximize the information gain, which has been shown to be an NP-hard problem \cite{2008-Krause-NearOptimal}.     

Sparse GPs use a small and special subset of data, namely inducing variables, and apply two key ideas: an assumption of conditional independence between training data and testing data given inducing variables; and the approximated conditional distribution for training data given inducing variables, for example, the fully or partially independent training conditional (FITC or PITC) \cite{2005-Quinonero-Candela-AUnifyingViewOfSparseGP,2006-Snelson-SGPUsingPseudoInputs,2007-Snelson-LocalAndGlobalSGP, 2015-Biji-OnlineFITCPITC}. Based on these assumptions, sparse GPs reduce the complexity to $\mathcal{O}(NM^2)$ where $M \ll N$ is the number of inducing variables. By contrast, the product of experts (PoEs) or Bayesian committee machines (BCM) \cite{2000-Tresp-BayesianCommitte, 2014-Cao-GeneralizedPoEs,2015-deisenroth-distributedGP} does not require to find a special set of variables. Those methods divide the large training data into segments and uses each with the GP model as a `local expert'. The final result are formed by multiplying the results given by each local experts, usually with certain weights in order to be robust against outliers. 

Several important questions remain for scalable GP methods. First, how are these different methods connected to each other? Second, how do we quantify the difference between an approximated GP posterior distribution and a full GP posterior distribution? This work aims to address both questions. First, we show that various approximated GP posteriors can be derived by applying a general belief updating framework \cite{2016-Bissiri-UpdatingBelief}. Using composite likelihoods \cite{2011-Varin-CompositeLikelihood}, we obtain a scalable composite GP model, which can be implemented in a recursive fashion. Second, we analyze how the posterior of the composite GPs differs from that of a full GP with respect to predictive performance and its representation of uncertainty \cite{1979-Bernardo-ReferencePosterior,1991-Cover-InfoTheory}.

This paper is organized as follows: In Section \ref{sec:method}, we derive the composite GP posterior. In Section \ref{sec:computation}, we give equations for scalable GP learning and prediction. The performance analysis of composite GPs is presented in Section \ref{sec:analysis}. Examples and discussions are presented in Section \ref{sec:example} and \ref{sec:conclusion}.


\section{Problem Formulation}\label{sec:problem}
We consider $f(\x)$ in \eqref{eq:model} to be a stochastic process modeled as a Gaussian process $\mathcal{GP}( \mu(\x), \sigma(\x,\x'))$ \cite{2006-Rasmussen-GPML}. This model yields a prior belief distribution $p(\z)$ over the latent variable at $M$ tests points:
\begin{equation}
\z = [f(\x^\star_1) , \cdots, f(\x^\star_M)]^\T \: \sim \: \mathcal{N}( \mvec_\z, \covmat_\z ).
\label{eq:latentvariable}
\end{equation}
The joint mean $\mvec_\z$ and covariance matrix $\covmat_\z$ are functions of the test points $\{ \x^\star_1, \dots,  \x^\star_M \}$. Using a training data set
$$\dataset = \big\{ (\x_1, y_1), \dots, (\x_N, y_N)  \big\} = \{ \X, \y \},$$
our goal is to update the belief distribution over $\z$ so as to produce a prediction $\what{\z}$ with a dispersion measure for uncertainty.

In addition to \eqref{eq:latentvariable}, we model $\eps_i$ in each sample from \eqref{eq:model} as
\begin{equation}
\eps_i \sim \mathcal{N}(0, \epsvar) \quad  \text{(i.i.d.)}
\end{equation}
This specifies a conditional data model $p(\y| \z)$, that is Gaussian with mean $\mvec_{\y|\z}$ and covariance $\covmat_{\y|\z}$. The standard Bayesian inference framework then updates the prior belief distribution $p(\z)$ into the posterior $p(\z | \y)$ via Bayes' rule. 

The inference of $\z$ depends, moreover, on a specified mean and covariance model, which can be learned in several ways. When it is parameterized  by a vector $\pvec$, the maximum marginal likelihood approach is a popular learning approach that aims to solve the problem
\begin{equation}
\what{\pvec} = \underset{\pvec}{\arg\max} \: \int p_\psub(\y|\z)p_\psub(\z) d\z,
\label{eq:MML}
\end{equation}
which requires several matrix inversions in itself. Finally, the posterior 
\begin{equation} 
p_{\psub}(\z | \y) = \frac{p_\psub(\y| \z)p_\psub(\z)}{p_\psub(\y)}
\label{eq:posterior}
\end{equation}
is evaluated at $\pvec = \what{\pvec}$. Both \eqref{eq:MML} and \eqref{eq:posterior} require a runtime on the order $\mathcal{O}(N^3)$ and a storage that scales as $\mathcal{O}(N^2)$, which renders standard GP training and inference intractable for large $N$ \cite{2005-Quinonero-Candela-AUnifyingViewOfSparseGP}.

In this work, our goal is to firstly formulate an alternative update of the belief distribution, secondly provide a scalable training and inference method and finally present an analysis on the performance of approximation in terms of MSE and information loss. 


\section{Updating Belief Distributions}\label{sec:method}

The posterior above can be thought of as a special case of updating the belief distribution $p(\z)$ into a new distribution $q(\z)$ using the data $\dataset$. A more general belief updating framework was formulated in \cite{2016-Bissiri-UpdatingBelief}. Using this framework, we first define a loss function $\loss(\y; \z)$ and then find the distribution $q(\z)$ which minimizes the average loss
\begin{equation}
L\big(q(\z)\big) \triangleq  \int_{\mathcal{Z}}  \loss(\y ; \z) q(\z) \: d\z + D(q(\z) || p(\z)),
\label{eq:lossFunction}
\end{equation}
regularized by the Kullback--Leibler divergence (KLD) $D(q(\z) || p(\z))$ \cite{1991-Cover-InfoTheory}. The first term in \eqref{eq:lossFunction} fits $q(\z)$ to the data, while the second term constrains it to the prior belief distribution. The updated belief distribution is then obtained as
\begin{equation}
	\widehat{q}(\z) = \underset{q(\z)}{\arg\min}\ L\big(q(\z)\big).
	\label{eq:minLoss}
\end{equation}
It is readily seen that for the loss function
\begin{equation}
\loss_{\text{GP}}(\y;\z) = - \ln p(\y|\z)
\label{eq:GPloss}
\end{equation}
the minimizer of $L_{\text{GP}}(q(\z))$ is the posterior $p(\z | \y)$, cf. \cite{2016-Bissiri-UpdatingBelief} for more details. It is also interesting to point out that the above optimal belief updating framework has the same structure as the KLD optimization problem in variational Bayesian inferences \cite{2012-Fox-VariationalBayesian}. Here, however, the problem is considered from a different angle: in \cite{2012-Fox-VariationalBayesian}, the challenge is to tackle the joint distribution of high-dimensional $\z$ by factorizing it into single variable factors; in this paper the challenge is to tackle the distribution of large data sets $\y$.

To alleviate the computational requirements using the GP loss \eqref{eq:GPloss}, we formulate a different loss function based on marginal blocks of the full data distribution $p(\y| \z)$ similar to the composite likelihood approach, cf. \cite{2011-Varin-CompositeLikelihood}. Specifically, we choose
\begin{equation}
\loss_{\text{CGP}}(\y ; \z) = - \sum_{k=1}^{K} \ln p(\y_{k} | \z),
\label{eq:compositeLikelihood}
\end{equation}
where the data set $\mathcal{D}$ has been divided into $K$ segments
\begin{equation}
\mathcal{D}_k = \{\X_k, \y_k\}, \; k=1,\dots, K,
\end{equation}
with $N_k \ll N$ samples each. As we show below, this cost function enables scalable and online processing for large data set. 

\begin{theorem}[Composite GP (CGP) update]\label{theorem:posteriors}
By applying \eqref{eq:compositeLikelihood}, we obtain a recursively updated belief distribution
\begin{equation}
\begin{split}
p_{\text{CGP}}(\z | \y_{1:K}) &\triangleq \underset{q(\z)}{\arg\min}\ L_{\text{CGP}}\big(q(\z) \big) \\
&= \frac{p_{\text{CGP}}(\z | \y_{1:K-1})p(\y_{K} | \z)}{p(\y_{K})},
\end{split}
\label{eq:cgpPosterior}
\end{equation} 
where $$p(\y_{1:K}) = \int_{\mathcal{Z}} p(\z)\prod_{k=1}^{K}p(\y_{k} | \z) d\z$$ is the marginalized distribution for all data; and $$p(\y_K) = \int_{\mathcal{Z}} p_{\text{CGP}}(\z | \y_{1:K-1})p(\y_{K} | \z) d\z$$ is the marginalized distribution for segment $K$. For a fixed $N_k$, \eqref{eq:cgpPosterior} can be evaluated in the runtime of order $\mathcal{O}(KN^3_k)$.
\end{theorem}

The proof follows by recognizing that \eqref{eq:lossFunction} is equivalent to the divergence
\begin{equation}
D\big( \: q(\z) \: || \: \exp(-\loss(\y; \z)  ) p(\z) \: \big) \: \geq \: 0,     
\end{equation}
which attains the minimum 0 only when $q(\z) \propto \exp(-\loss(\y; \z)  )p(\z)$. Then the result is obtained by noting that $q(\z)$ is a distribution that integrates to unity. The recursive computation of the CGP posterior in (\ref{eq:cgpPosterior}) can be tackled using standard tools \cite{2013-Srkk-BayesianFilteringAndSmoothing} as discussed in Section \ref{sec:computation} below.

Here it is instructive to compare the CGP with the SGP approach \cite{2006-Snelson-SGPUsingPseudoInputs} which is formulated using a set of latent `inducing variables' $\indvar$ with a joint distribution $p(\y, \indvar|\z)$. By assuming that $\y$ is independent of $\z$ when given $\indvar$, the joint distribution can be factorized into $p(\y|\indvar)p(\indvar|\z)$. Then the implicit loss function used in SGP is based on marginalizing out $\indvar$ from the joint distribution:
\begin{equation}
\begin{split}
\loss_{\text{SGP}}(\y ; \z) &=  - \ln \int_{\mathcal{U}}p(\y|\indvar) p(\indvar | \z) d\indvar,
\end{split}
\label{eq:sparseLikelihood}
\end{equation}
using a segmented model $p(\y|\indvar) = \prod_{b=1}^{B}p(\y_{b} | \indvar)$. The SGP posterior is 
\begin{equation}
\begin{split}
p_{\text{SGP}}(\z | \y) &\triangleq \underset{q(\z)}{\arg\min}\ L_{\text{SGP}}\big( q(\z) \big) \\
&= \frac{p(\z)\int_{\mathcal{U}} \prod_{b=1}^{B}p(\y_{b} | \indvar) p(\bm{u} | \z) d\bm{u}}{\int_{\mathcal{U}} \widetilde{p}(\y | \bm{u})p(\bm{u}) d\bm{u}}.
\end{split}
\label{eq:sgpPosterior}
\end{equation}
This formulation reproduces the FITC and PITC approximations depending on the size of the segments $\y_{b}$, cf. \cite{2005-Quinonero-Candela-AUnifyingViewOfSparseGP, 2007-Snelson-LocalAndGlobalSGP}. The inducing variables are targeting a compressed representation of the data and the information is transferred to the belief distribution of $\z$ via $p(\indvar | \z)$. Thus $\indvar$ must be carefully selected so as to transfer the maximum amount of information about the training data $\y$. The optimal placement of inducing variables involves a challenging combinatorial optimization problem and but can be tackled using greedy search heuristics \cite{2003-Seeger-FastForwardSelection,2008-Krause-NearOptimal,2009-Titsias-VariationalLearning}. 


\section{Recursive Computation}\label{sec:computation}
The model parameters $\pvec$ are fixed unknown quantities and typically learned using the maximum likelihood (ML) method. Once this is completed, the prediction of $\z$ along with its dispersion is computed. In this section, we present recursive computations for both CGP learning and prediction. 

\subsection{Learning}
\begin{figure*}[ht]
	\centering
	\subfigure[$N_k = 50$, $k \in 1, \dots, 100$]{\includegraphics[width = .225\linewidth]{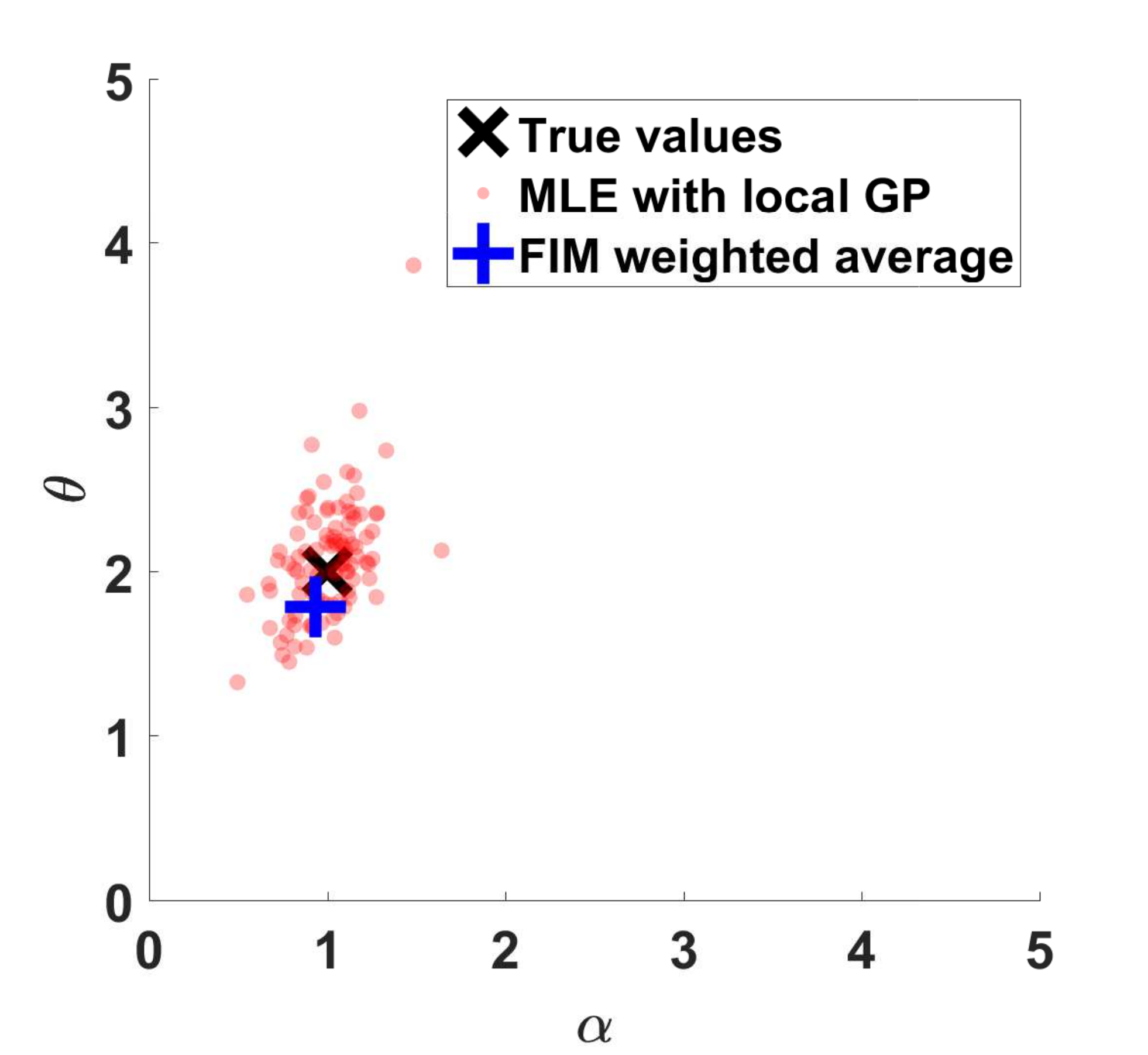}}\qquad 
	\subfigure[$N_k = 100$, $k \in 1, \dots, 50$]{\includegraphics[width = .225\linewidth]{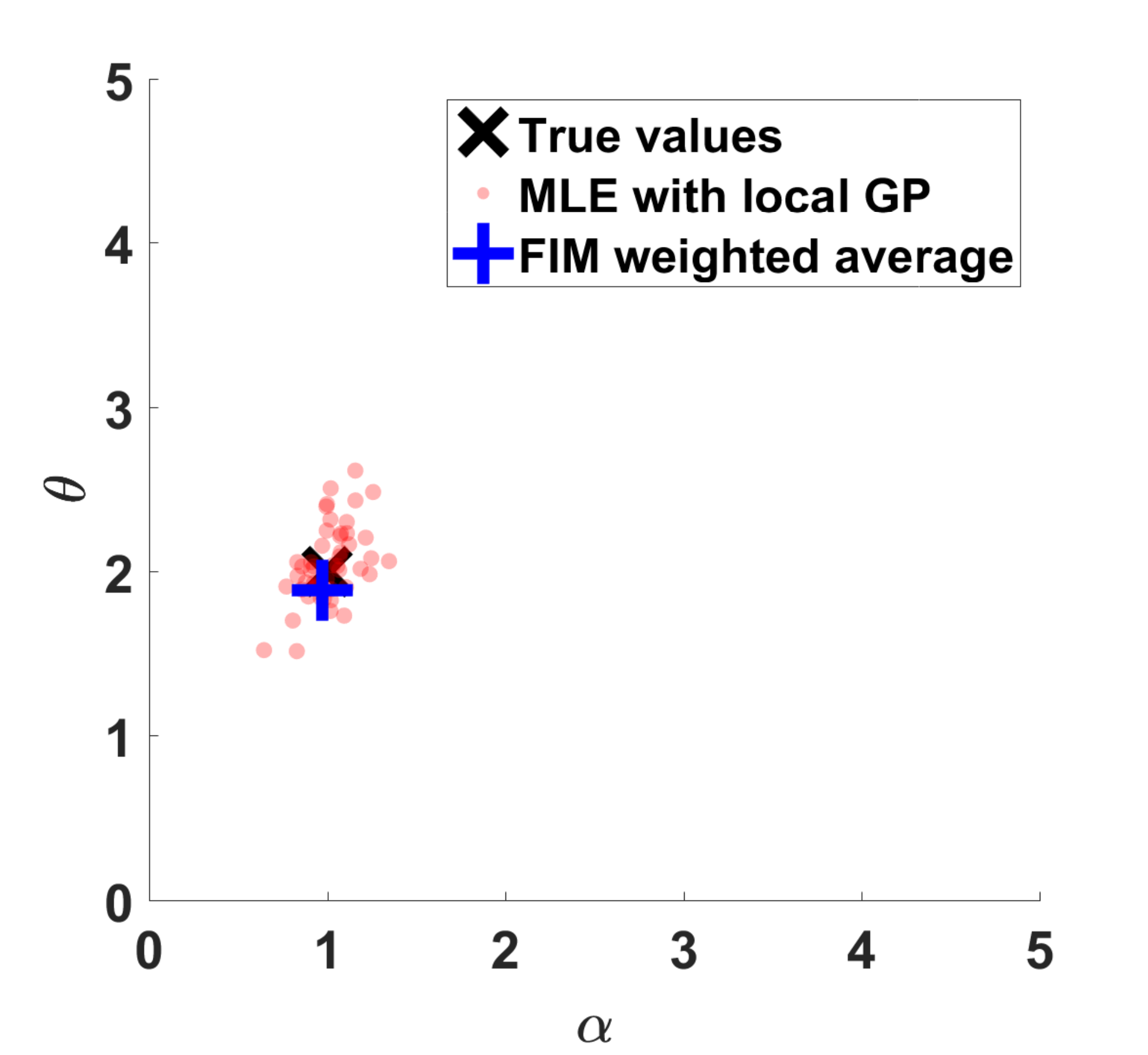}}\qquad
	\subfigure[$N_k = 200$, $k \in 1, \dots, 25$]{\includegraphics[width = .225\linewidth]{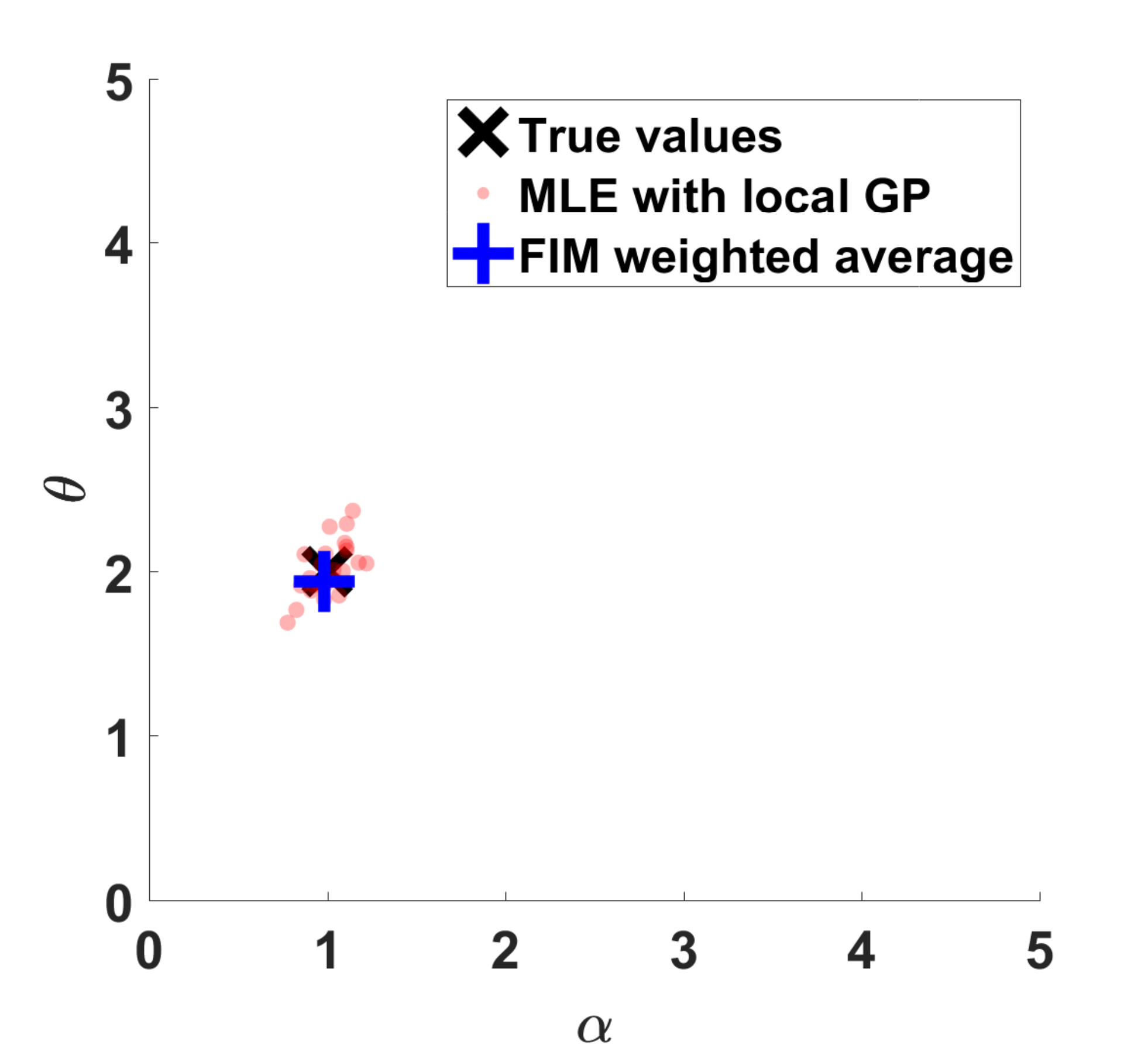}}\qquad
	\caption{FIM weighted average of MLEs based on segmented training data with different block length $N_k$.}
	\label{fig:example_fim}
\end{figure*}

The Fisher information matrix (FIM), $\bm{J}(\bm{\theta} | \y_k)$, quantifies the information about the model parameters $\bm{\theta}$ contained in data $\y_k$ when assuming a model $p_{\psub}(\y_k)$. In case of Gaussian distributed data, the FIM can be obtained by the Slepian-Bangs formula. We refer readers to (B.3.3) in \cite{1997-Stocia-SpectralAnalysis} for this formula. The larger the FIM in the L\"{o}wner order sense, the lower errors we may achieve when learning the optimal model parameters. Indeed, the inverse of the FIM, provides an estimate of the variance of an efficient estimator $\what{\pvec}$ \cite{1993-Kay-StatisticalSignalProcessing,2013-VanTrees-DetectionEstimation}.

Each data segment $\y_k$ yields its own maximum likelihood estimate $\what{\pvec}_k$ with an approximate error covariance matrix $\what{\bm{J}}_k = \bm{J}(\what{\pvec}_k | \y_k)$. When the segments are obtained from the same data generating process, the model is applicable to each segment and we combine the estimates to provide a refined model parameter at segment $K$:
\begin{equation}	
\bar{\bm{\theta}}_K = \left(\sum_{k=1}^{K} \what{\bm{J}}_k\right)^{-1}\left(\sum_{k=1}^{K} \what{\bm{J}}_k \widehat{\pvec}_k \right) = {\bm{\Lambda}}^{-1}_K\bm{s}_K
\label{eq:MLcombination}
\end{equation}
The quantities are computed recursively as
\begin{align}
{\bm{\Lambda}}_k &= {\bm{\Lambda}}_{k-1} + \what{\bm{J}}_k, \\
\bm{s}_k &= \bm{s}_{k-1} + \what{\bm{J}}_k\widehat{\pvec}_k,
\end{align}
where $\bm{\Lambda}_{0} = \bm{0}$ and $\bm{s}_{0} = \bm{0}$, cf. \cite{2017-Zachariah-ClockSynchronization}.

Figure \ref{fig:example_fim} illustrates the recursive learning of a CGP learning for a model with zero-mean function and the squared exponential (SE) covariance function ($\bm{\theta} = [1, 2]^\top$) is shown in Figure \ref{fig:example_fim}. In the example we split the training data ($N = 5000$) into a number of segments. Three cases with different segmented data length ($N_k = 50$, $100$, and $200$) are shown. Note that the ML estimator is asymptotically efficient estimator and thus the accuracy of $\what{\pvec}_k$ and $\what{\bm{J}}_k$ improves as $N_k$ becomes larger. Therefore there is a trade-off between accuracy and computational cost. Setting $N_k = 200$, the accuracy of the weighted combination \eqref{eq:MLcombination} is satisfying, while the computational complexity is still maintained at a relatively low level comparing to the full GP learning.

It is worth pointing out that, the above idea of segmenting data and weighting the contributions from each module is similar to the approach found in the recent work by Jacob, \etal \yrcite{2017-Jacob-BetterTogether?}. The modularized inference increases robustness to misspecifications of the joint data model.

\subsection{Prediction}
We now present the mean and covariance of the updated belief distribution for CGP, cf. Theorem \ref{theorem:posteriors}. Specifically, given the CGP posterior distribution $p_{\text{CGP}}(\z | \y_{1:k-1}) = \mathcal{N}(\widetilde{\mvec}_{\z|\y_{1:k-1}}, \widetilde{\covmat}_{\z|\y_{1:k-1}})$ and a new data segment $\y_{k}$, the posterior $p_{\text{CGP}}(\z | \y_{1:k}) = \mathcal{N}(\widetilde{\mvec}_{\z|\y_{1:k}}, \widetilde{\covmat}_{\z|\y_{1:k}})$ is updated recursively, cf. \cite{2013-Srkk-BayesianFilteringAndSmoothing}. 

First, a prior is constructed. For $k = 1$, the GP prior in \eqref{eq:latentvariable} is used as the prior distribution for $\z$; for $k \geq 2$, previous posterior $p_{\text{CGP}}(\z | \y_{1:k-1})$ is used as the new prior. 

Second, the conditional distribution $p(\y_{k} | \z)$ is obtained with mean and covariance matrix
\begin{align}
\mvec_{\y_k|\z} &= \bm{\mu}_{\y_k} + \covmat_{\y_k,\z}\covmat^{-1}_{\z,\z}(\mvec_{\z|\y_{1:k-1}} - \mvec_{\z}), \\
\covmat_{\y_k|\z} &= \covmat_{\y_{k, k}} - \covmat_{\y_k, \z}\covmat^{-1}_{\z,\z}\covmat_{\z, \y_k}. 
\end{align}

Finally, the posterior $p_{\text{CGP}}(\z | \y_{1:k})$ has a mean and covariance:
\begin{align}
\begin{split}
\widetilde{\mvec}_{\z|\y_{1:k}} =& \widetilde{\mvec}_{\z|\y_{1:k-1}} + \widetilde{\covmat}_{\z|\y_{1:k-1}}\bm{H}^{\top}_{k}\bm{G}_k^{-1}\\
&\: \: \: [\y_k - \bm{\mu}_{\y_k} - \bm{H}_k(\widetilde{\mvec}_{\z|\y_{1:k-1}} - \bm{\mu}_{\z})], 
\end{split} \\
\begin{split}
\widetilde{\covmat}_{\z|\y_{1:k}} =& \widetilde{\covmat}_{\z|\y_{1:k-1}} - \widetilde{\covmat}_{\z|\y_{1:k-1}}\bm{H}^{\top}_{k}\bm{G}_k^{-1}\bm{H}_{k}\widetilde{\covmat}_{\z|\y_{1:k-1}}.
\end{split}
\end{align}
The matrices $\bm{H}_k$ and $\bm{G}_k$ are defined as
\begin{align}
\bm{H}_k &\triangleq \covmat_{\y_k,\z}\covmat^{-1}_{\z,\z}, \\
\bm{G}_k &\triangleq \covmat_{\y_k|\z} + \bm{H}_{k}\widetilde{\covmat}_{\z|\y_{1:k-1}}\bm{H}^{\top}_{k}. 
\end{align}

Together the equations form a recursive computation where $p_{\text{CGP}}(\z | \y_{1:k})$ is the basis for obtaining  $p_{\text{CGP}}(\z | \y_{1:k+1})$ after observing $\y_{k+1}$.


\section{Performance Analysis of Composite GPs}\label{sec:analysis}
Given the appealing computational properties of CGP, a natural question is how good is its posterior as compared with that of GP? Specifically, we are interested in the predictive performance and the ability to represent uncertainty about the latent state $\z$, which will be addressed in the following subsections. Both aspects are related to the amount of information that the training data provides about the latent state, i.e., $I(\z; \y)$ \cite{1991-Cover-InfoTheory}.


\subsection{Data-Averaged MSE}
We begin by considering an arbitrary test point, so that $z$ is scalar and $\what{z}$ is any predictor. Given the prior belief distribution of $z$ with known variance $\sigma^2_{z}$, the data-averaged mean squared error (MSE) is lower bounded by the mutual information between $z$ and the training data $\y$:
\begin{equation}
\mathbb{E}[(z - \widehat{z})^2] \; \geq \; \sigma^2_{z} \: \frac{1}{2\pi e}\exp[-2I(z; \y)],     
\label{eq:mseBound}
\end{equation}
under fairly general conditions, cf. Theorem 17.3.2 in \cite{1991-Cover-InfoTheory}. When the marginal data distribution $p(\y)$ obtained from GP is well-specified, the bound \eqref{eq:mseBound} equals the posterior variance $\sigma^2_{z|\y}$ of the GP and is attained by setting $\what{z} = \mu_{z|\y}$. Thus $I(z; \y)$ also represents the reduced uncertainty of the updated belief distribution for GP.




In this scenario, what is the additional MSE incurred when using the CGP posterior mean as a predictor $\what{z}_{\text{CGP}}$? In general, the data-averaged MSE equals
\begin{equation}
\mathbb{E}[(z - \what{z}_{\text{CGP}})^2] 
= {\sigma}^2_{z|\y} + \mathbb{E}[(\mu_{z|\y} - \wtilde{\mu}_{z|\y_{1:K}})^2],
\label{eq:mse}
\end{equation}
where the second term is the additional MSE given by the difference between the GP and CGP posterior means, respectively. To obtain closed-form expressions of this term, we consider the case of $K=2$ segments.

\begin{theorem}[Excess MSE of CGP]\label{theorem:mse}
The additional MSE of CGP when $K = 2$ equals
\begin{equation}
\begin{split}
    \mathbb{E}[(\mu_{z|\y} - \wtilde{\mu}_{z|\y_{1:K}})^2] =
    \begin{bmatrix}
    \bm{\alpha}_1 \\
    \bm{\alpha}_2
    \end{bmatrix}^\top
    \begin{bmatrix}
    \bm{\Sigma}_{\y_{1, 1}} & \bm{\Sigma}_{\y_{1, 2}}\\
    \bm{\Sigma}_{\y_{2, 1}} & \bm{\Sigma}_{\y_{2, 2}}
    \end{bmatrix}
    \begin{bmatrix}
    \bm{\alpha}_1\\
    \bm{\alpha}_2
    \end{bmatrix},
\end{split}
\label{eq:dataAveragedMSE}
\end{equation}
where $\bm{\Sigma}_{\y_{i, j}}$ is the covariance matrix of the $i$-th and $j$-th data segments.
The vectors $\bm{\alpha}_1$ and $\bm{\alpha}_2$ are given by
\begin{equation}
\begin{split}
    \begin{bmatrix}
    \bm{\alpha}_1\\
    \bm{\alpha}_2
    \end{bmatrix} =& \left(
    \begin{bmatrix}
    A & B \\
    C & D
    \end{bmatrix} - \begin{bmatrix}
    A' & B' \\
    C' & D'
    \end{bmatrix}\right)^\top\begin{bmatrix}
    \bm{\Sigma}_{z, \y_1} \\
    \bm{\Sigma}_{z, \y_2} 
    \end{bmatrix},
\end{split}
\end{equation}
where $\bm{\Sigma}_{z, \y_i}$ is the column vector of covariances between the testing data and the $i$-th data segment; and the two block matrices ($[A,\: B;\: C,\: D]$ and $[A',\: B';\: C',\: D']$)  are coefficient matrices for the full GP and CGP predictions, respectively.  
\end{theorem}

The full derivation is  omitted here due to page limitations. We present a sketch of the proof in the following. It can be shown that the coefficient matrix for GP prediction is the block inverse of the covariance matrix of data $\y$:
\begin{equation}
\begin{split}
    &\begin{bmatrix}
    A & B \\
    C & D
    \end{bmatrix} = \begin{bmatrix}
    \bm{\Sigma}_{\y_{1, 1}} & \bm{\Sigma}_{\y_{1, 2}}\\
    \bm{\Sigma}_{\y_{2, 1}} & \bm{\Sigma}_{\y_{2, 2}}
    \end{bmatrix}^{-1} = \\
    &\begin{bmatrix}
    \bm{\Sigma}^{-1}_{\y_{1, 1}}\bm{\Sigma}_{\y_{1, 2}}\bm{\Sigma}_{\y_{2}|\y_{1}}^{-1}\bm{\Sigma}_{\y_{2, 1}}\bm{\Sigma}^{-1}_{\y_{1, 1}} & -\bm{\Sigma}^{-1}_{\y_{1, 1}}\bm{\Sigma}_{\y_{1, 2}}\bm{\Sigma}_{\y_{2|1}}^{-1}\\
    -\bm{\Sigma}_{\y_{2|1}}^{-1}\bm{\Sigma}_{\y_{2, 1}}\bm{\Sigma}^{-1}_{\y_{1, 1}} & \bm{\Sigma}_{\y_{2|1}}^{-1}
    \end{bmatrix}
\end{split}
\end{equation}
where
\begin{equation}
    \bm{\Sigma}_{\y_{2|1}} = \bm{\Sigma}_{\y_{2,2}} - \bm{\Sigma}_{\y_{2,1}}\bm{\Sigma}^{-1}_{\y_{1,1}}\bm{\Sigma}_{\y_{1,2}}. 
\end{equation}

To derive the corresponding coefficient matrix for CGP, we first define the approximated covariance matrix between data block $i$ and $j$:
\begin{equation}
    \widetilde{\bm{\Sigma}}_{\y_{i,j}} \triangleq \bm{\Sigma}_{\y_i, z}\frac{1}{\sigma^2_{z}}\bm{\Sigma}_{z, \y_j},
\label{eq:testingDataAsInducingVar}
\end{equation}
which means the segmented observations are indirectly connected by the testing data. Thereafter, the approximated conditional covariance matrix of $\y_2$ given $\y_1$ can be expressed as  
\begin{equation}
    \widetilde{\bm{\Sigma}}_{\y_{2|1}} \triangleq \bm{\Sigma}_{\y_{2,2}} - \widetilde{\bm{\Sigma}}_{\y_{2,1}}\bm{\Sigma}_{\y_{1,1}}^{-1}\widetilde{\bm{\Sigma}}_{\y_{1,2}}. 
\end{equation}
After a few steps of algebraic manipulation, the block coefficients matrix for CGP can be obtained as
\begin{equation}
    \begin{bmatrix}
    A' & B' \\
    C' & D'
    \end{bmatrix} = \begin{bmatrix}
    \bm{\Sigma}^{-1}_{\y_{1,1}} & 0\\
    -\frac{{\sigma}^2_{z|\y_{1}}}{\sigma^2_{z}}\widetilde{\bm{\Sigma}}_{\y_{2|1}}^{-1}\widetilde{\bm{\Sigma}}_{\y_{2,1}}\bm{\Sigma}_{\y_{1,1}}^{-1} & \frac{{\sigma}^2_{z|\y_{1}}}{\sigma^2_{z}}\widetilde{\bm{\Sigma}}_{\y_{2|1}}^{-1}
    \end{bmatrix}
\end{equation}
where ${\sigma}^2_{z|\y_{1}} = \sigma^2_{z} - \bm{\Sigma}_{z, \y_1}\bm{\Sigma}^{-1}_{\y_1, \y_1}\bm{\Sigma}_{\y_1, z}$ is the posterior variance of latent variable $\z$ using the first data block $\y_1$. 

By comparing the expressions for the coefficients matrices of GP and CGP ($K = 2$), we can make the following observations. First, for CGP, the covariance matrix $\bm{\Sigma}_{\y_{2,1}}$ is replaced by an approximation $\wtilde{\bm{\Sigma}}_{\y_{2,1}}$, which relies on the testing points to `connect' the two blocks, cf. \eqref{eq:testingDataAsInducingVar}. Second, the CGP prediction effectively assumes that the segmented training data blocks are conditionally independent, when given testing data. Therefore, the coefficients matrix $A'$ is simply the inverse of first data block's covariance matrix, and $B' = 0$. Thus Theorem \ref{theorem:mse} provides a means of probing the sources of the additional MSE incurred when using CGP compared to GP.

\subsection{Data-Averaged KLD}

Another way to compare different updated belief distributions is to quantify how much they differ from the prior distribution $p(\z)$. Specifically, we use the Kullback--Leibler divergence $D( p(\z | \y) || p(\z))$ and average it over all realizations $\y$ under the marginal data model. For GP, it is straight-forward to show the identity $\mathbb{E}_{\y}[D( p(\z | \y) || p(\z)) ]= I(\z ; \y)$, which is the average information gain about $\z$ provided by the data and the selected model, cf. \cite{1979-Bernardo-ReferencePosterior}.



\begin{theorem}[Difference between average information gains]\label{theorem:dataAveKLD}
The data-averaged KL divergences of the posterior belief distributions differ by
\begin{equation}
\begin{split}
&\mathbb{E}_{\y}[D(p(\z | \y) || p(\z))] - \mathbb{E}_{\y_{1:K}}[D(p_{\text{CGP}}(\z | \y_{1:K}) || p(\z))] \\
=& I(\z ; \y) - \sum_{k=1}^{K} I(\z; \y_{k}),
\end{split} 
\label{eq:dif_expected_info}
\end{equation}
where $I(\z ; \y)$ is the mutual information of the latent variable $\z$ and all observations $\y$, and $\sum_{k=1}^{K} I(\z; \y_{k})$ is the sum of mutual information of $\z$ and observation block $\y_k$.
\end{theorem}

\begin{figure*}[h]
	\centering
	\subfigure[GP posterior]{\includegraphics[width = .225\linewidth]{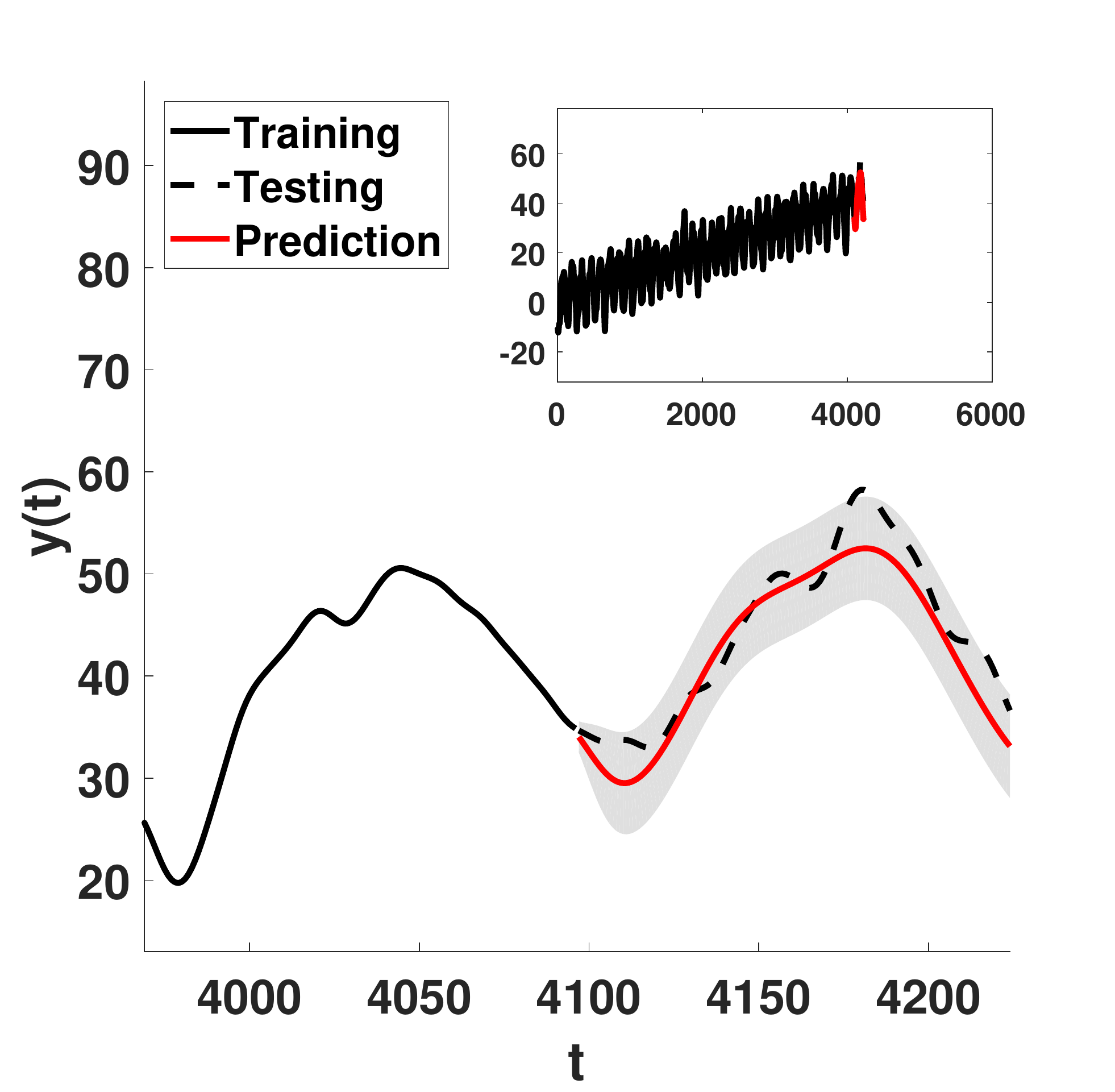}}\qquad
	\subfigure[CGP posterior]{\includegraphics[width = .225\linewidth]{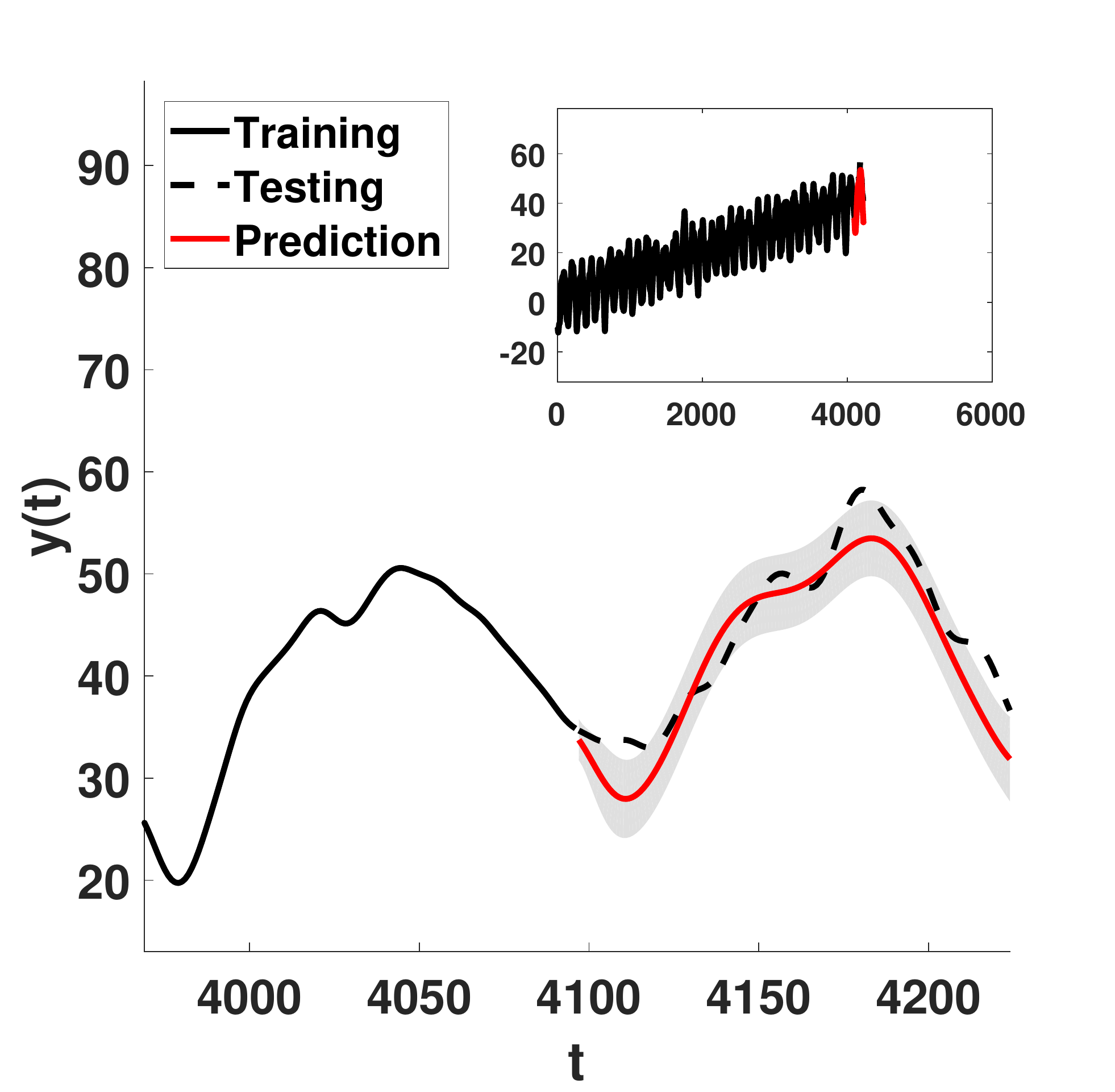}}\qquad
	\subfigure[SGP posterior]{\includegraphics[width = .225\linewidth]{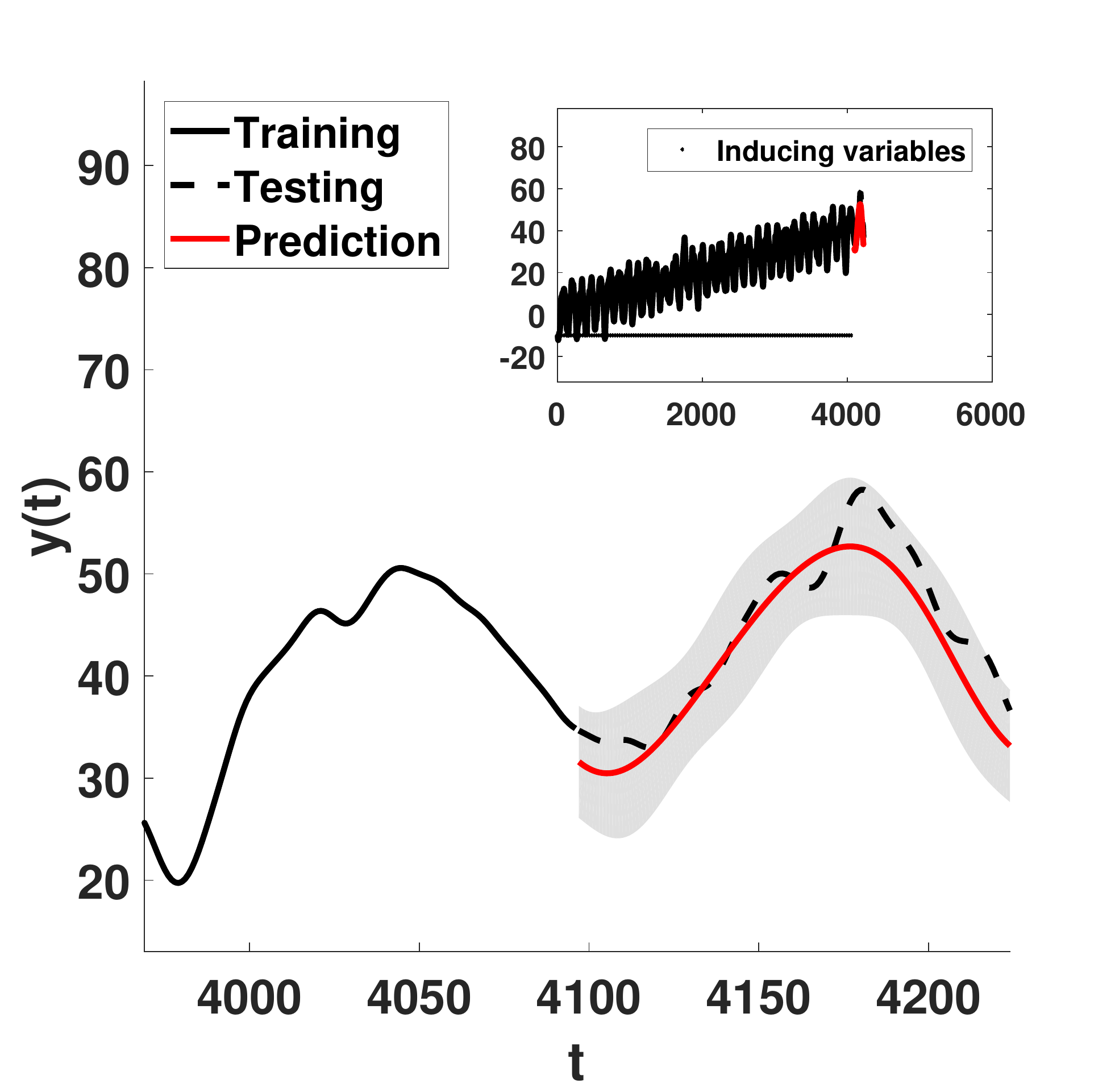}}\qquad \\
	\subfigure[GP covariance matrix]{\includegraphics[width = .225\linewidth]{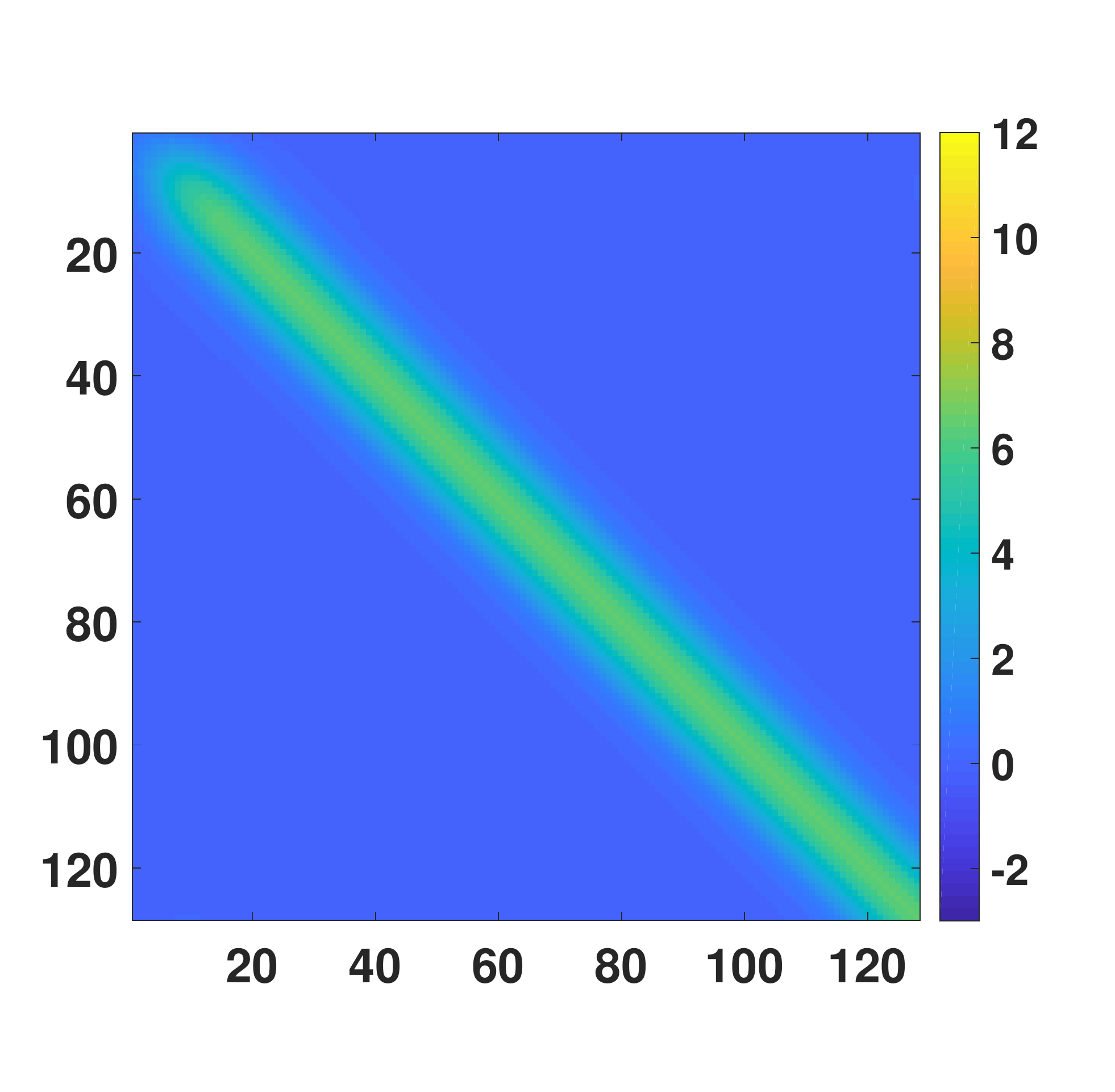}}\qquad
	\subfigure[CGP covariance matrix]{\includegraphics[width = .225\linewidth]{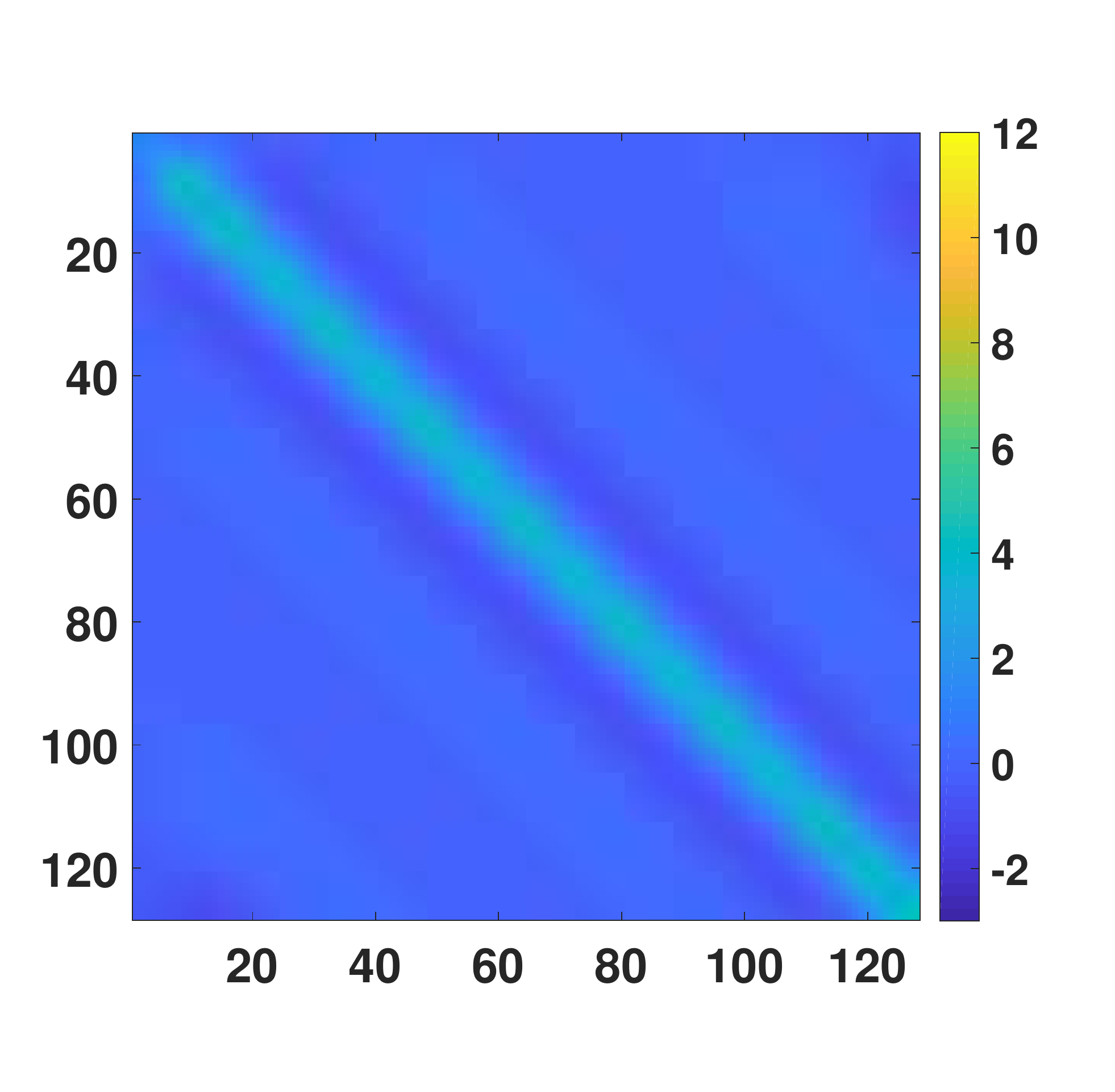}}\qquad
	\subfigure[SGP covariance matrix]{\includegraphics[width = .225\linewidth]{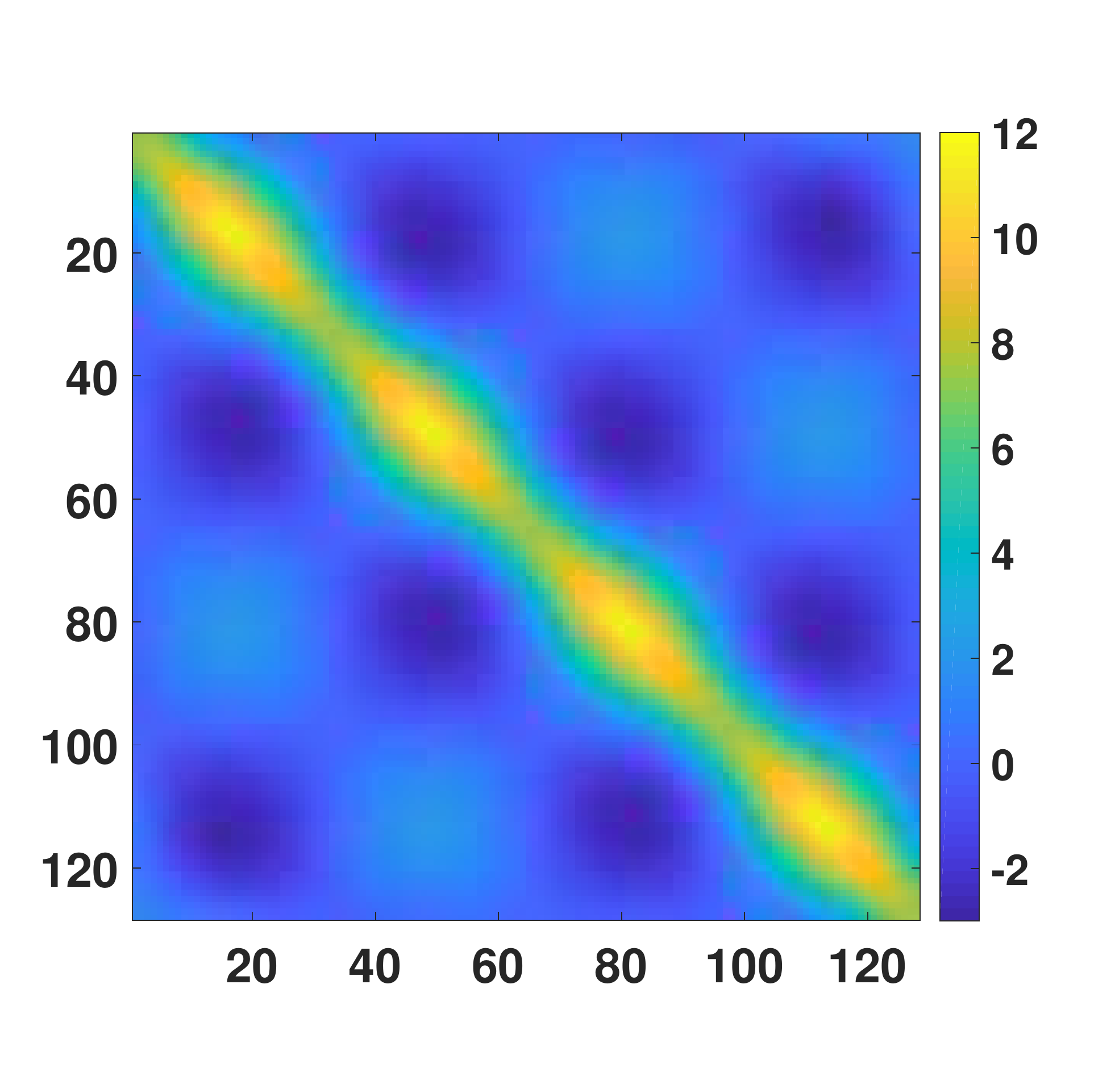}}\qquad
	\caption{The GP, CGP, and SGP predictive posterior distributions and covariance matrices.}
	\label{fig:syntheticData}
\end{figure*}

\begin{prf}
The data-averaged KLD between the CGP posterior and the prior is given by
\begin{equation}
\begin{split}
&\mathbb{E}_{\y_{1:K}}[D(p_{\text{CGP}}(\z | \y_{1:K}) || p(\z))] \\
=& \int_{\mathcal{Y}_{1:K}} p(\y_{1:K}) \int_{\mathcal{Z}} p_{\text{CGP}}(\z | \y_{1:K})\ln\frac{p_{\text{CGP}}(\z | \y_{1:K})}{p(\z)} d\z d\y_{1:K} \\
=& \sum_{k=1}^{K} \int_{\mathcal{Y}_k} \int_{\mathcal{Z}} p(\z , \y_k)[\ln\frac{1}{p(\z)} + \ln p(\z | \y_k)] d\z d\y_k \\
=& \sum_{k=1}^{K} I(\z; \y_k).
\end{split}
\end{equation}
Note that the second last holds because $p_{\text{CGP}}(\y_{1:K} | \z) = \prod_{k=1}^{K}p(\y_{k} | \z)$, $p(\y_{1:K}) = \int_{\mathcal{Z}} p(\z)\prod_{k=1}^{K}p(\y_{k} | \z) d\z$, and 
\begin{equation}
\begin{split}
&\int_{\mathcal{Y}_j} \int_{\mathcal{Y}_k} \int_{\mathcal{Z}} p(\y_{j} | \z) p(\z , \y_k) \ln\frac{p(\z | \y_k)}{p(\z)} d\z d\y_k d\y_j \\
=& \int_{\mathcal{Y}_k} \int_{\mathcal{Z}} p(\z , \y_k) \ln\frac{p(\z | \y_k)}{p(\z)} d\z d\y_k, \: \forall j \ne k.
\end{split}
\end{equation}
\end{prf}




\begin{remark}[Redundancy and synergy]\label{remark:synergyAndRedundancy}
The difference between information gains in Theorem \ref{theorem:dataAveKLD} can be positive or negative. If the difference is negative the segmented data is said to be \textbf{redundant} and there is overlapping information in the data segments about the latent variable; otherwise the segmented data are said to be \textbf{synergistic} in which case there is more information about the latent state by jointly considering all segmented data  \cite{2015-Barrett-InformationSharing,2014-SynergyRedundancy-Timme}.
\end{remark}



\begin{remark}[Under- and overestimation of uncertainty]\label{remark:overAndUnderestimate}
When the marginal data distribution obtained from GP is correctly specified, $I(z; \y)$ provides the correct measure of uncertainty about the latent variable, cf. \eqref{eq:mseBound}. In this scenario, the CGP belief distribution will either under- or overestimate the uncertainty depending on whether the data segments are redundant or synergistic in nature.
\end{remark}


\section{Examples}\label{sec:example}
In this section, we present examples of CGPs for processing large data sets. The first two examples are based on synthetic data. In the third example, we demonstrate the CGP for NOx predictions based on real-world data. 

\begin{figure*}[h]
	\centering
	\subfigure[GP mean]{\includegraphics[width = .225\linewidth]{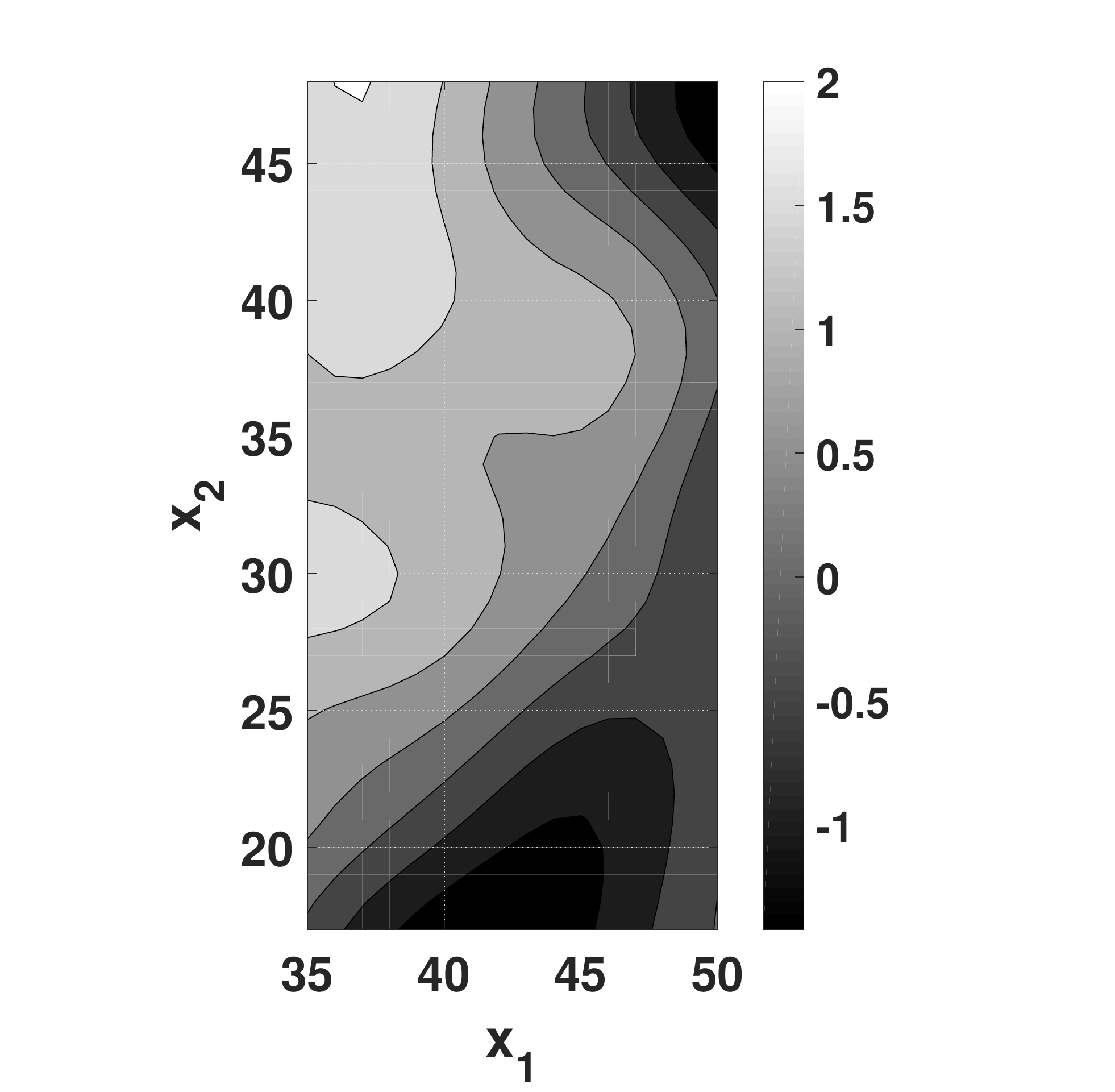}} 
	\subfigure[CGP mean, $K = 4$]{\includegraphics[width = .225\linewidth]{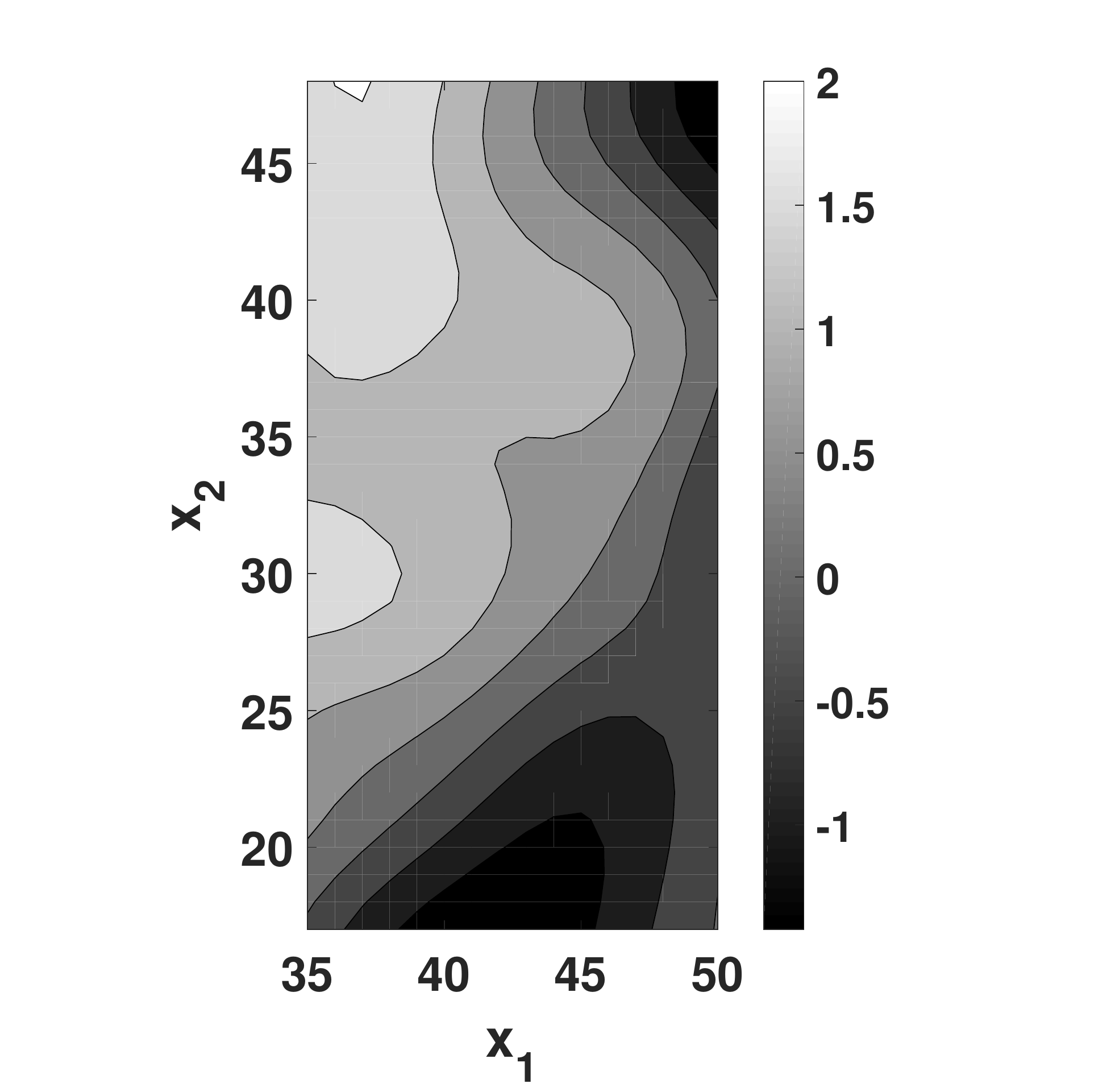}} 
	\subfigure[CGP mean, $K = 16$]{\includegraphics[width = .225\linewidth]{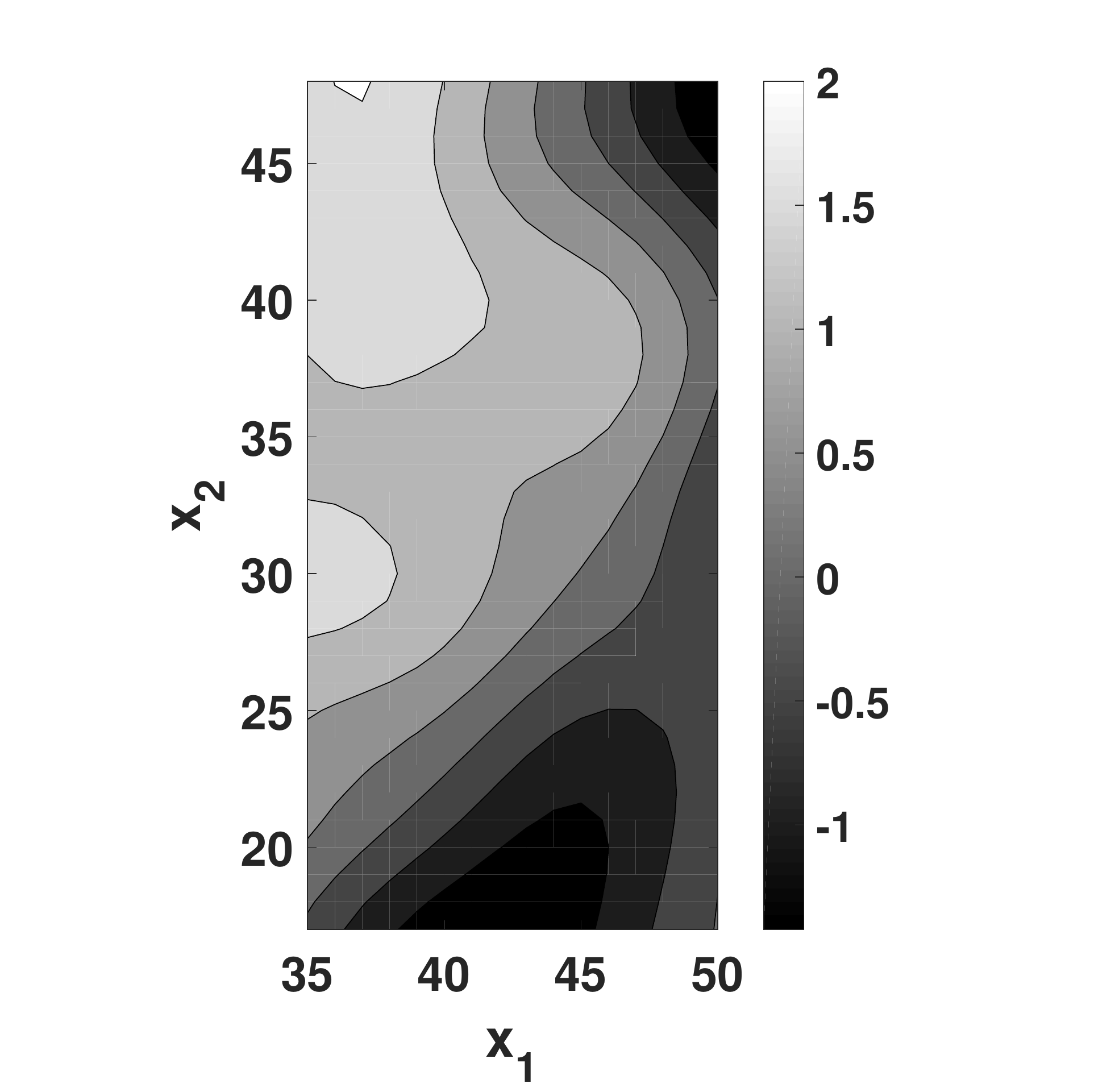}} 
	\subfigure[SGP mean]{\includegraphics[width = .225\linewidth]{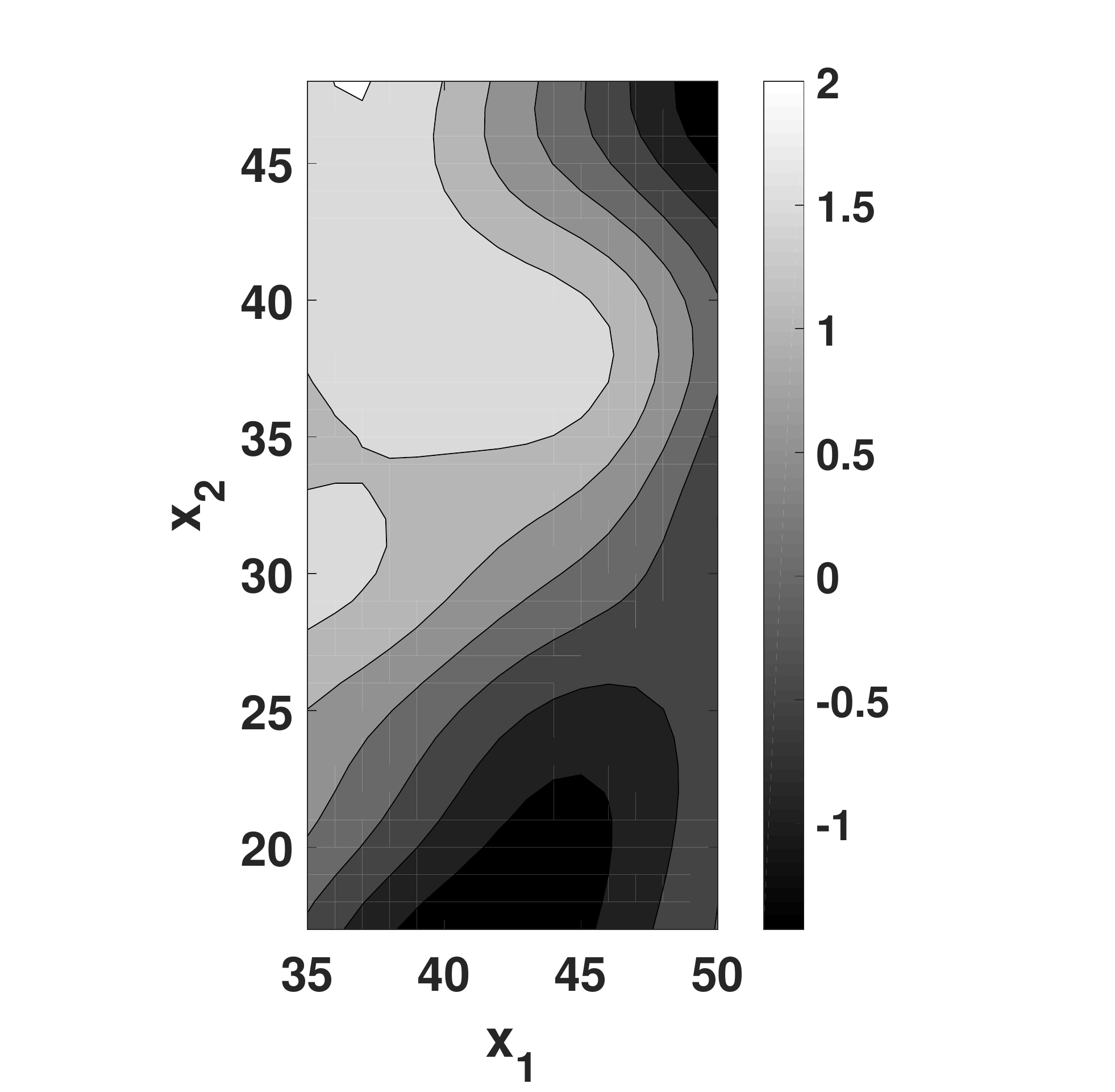}}  \\
	\subfigure[GP variance]{\includegraphics[width = .225\linewidth]{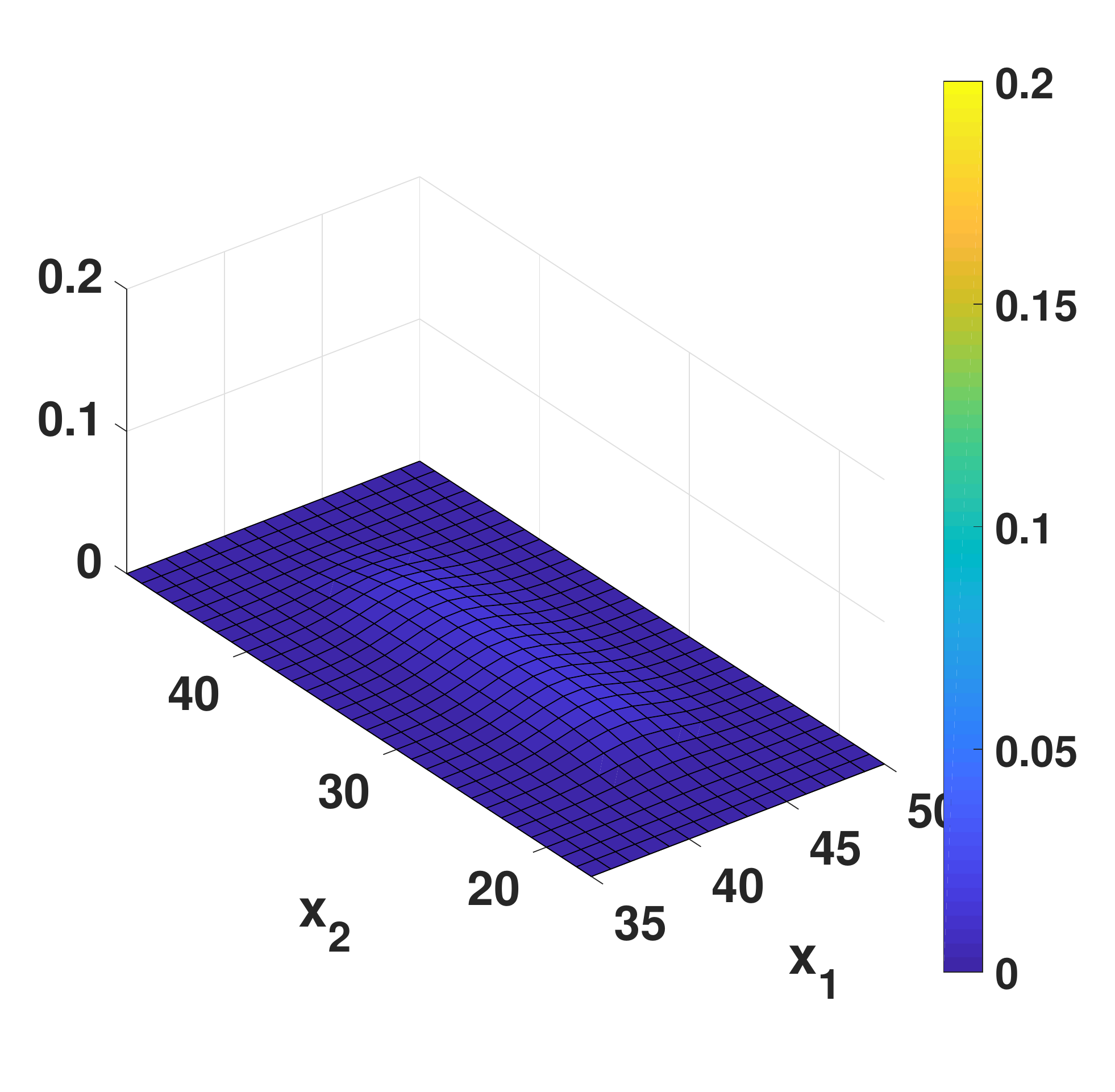}} 
	\subfigure[CGP variance, $K = 4$]{\includegraphics[width = .225\linewidth]{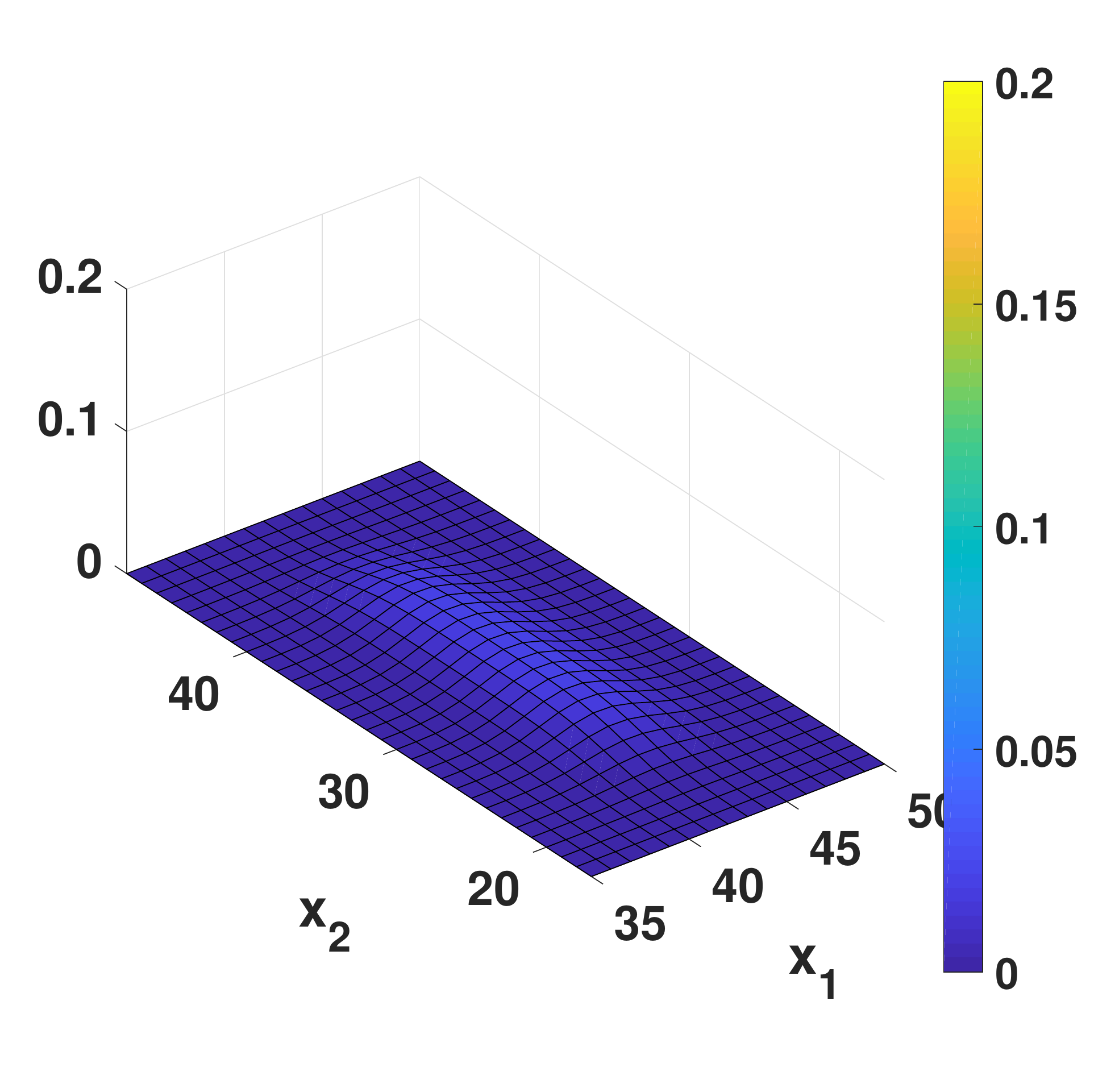}} 
	\subfigure[CGP variance, $K = 16$]{\includegraphics[width = .225\linewidth]{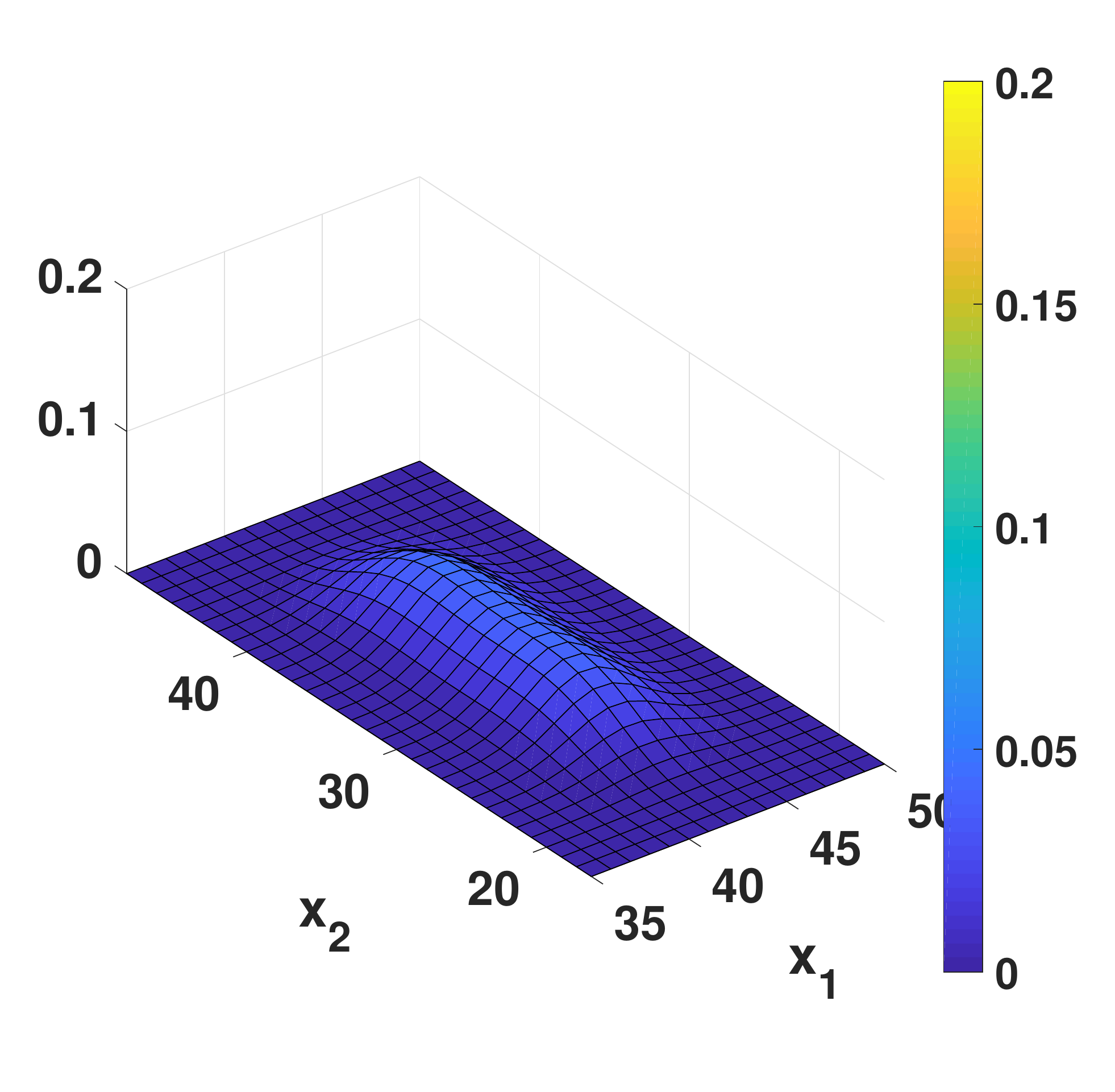}} 
	\subfigure[SGP variance]{\includegraphics[width = .225\linewidth]{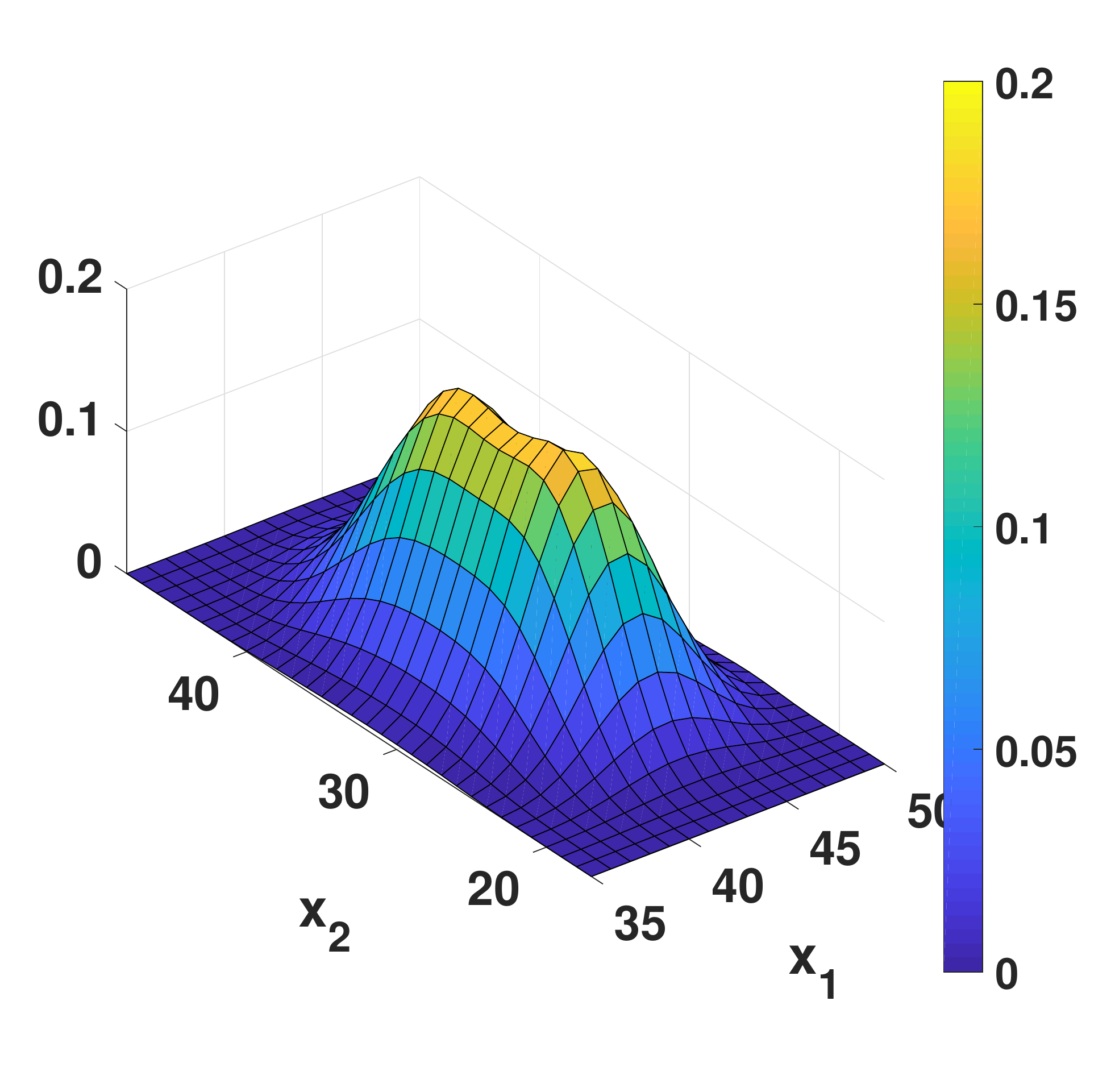}}
	\caption{The GP, CGP ($K = 4$ and $16$), and SGP interpolations for the missing values of GRF in Figure \ref{fig:grf}.}
	\label{fig:spatial}
\end{figure*}

\begin{figure}[ht]
	\centering
	\includegraphics[width=.8\linewidth]{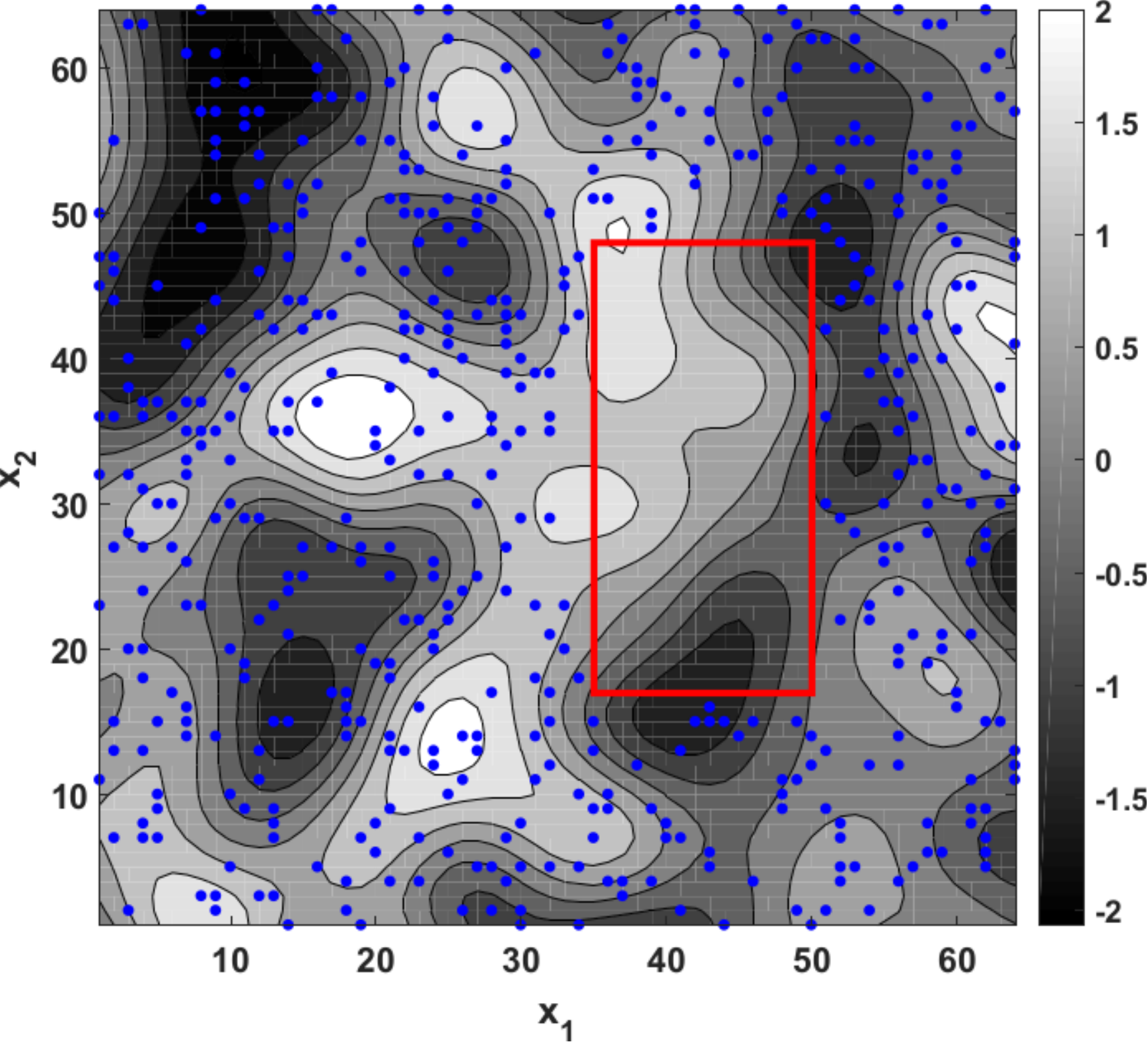}
	\caption{A realization of the GRF: points within the red box are used for testing; blue dots are the inducing variables.}
	\label{fig:grf}
\end{figure}

\subsection{Synthetic Time Series Data}
Considering the GP model with a linear mean function $\mu(t) = at+b$, and a covariance function with periodic patterns $\sigma(t, t') = \alpha_1^2\exp[\frac{-2\sin^2(\pi|t-t'|/T)}{\theta_1^2}] + \alpha_2^2\exp[\frac{-(t-t')^2}{\theta_2^2}] + \sigma^2_\epsilon$, where the period $T = 128$. In total 4224 data points are simulated: the first 4096 points are used as observations, and the last 128 points need to be predicted. Assuming the hyper-parameters are known, the GP, CGP, and SGP predictive posterior distributions are illustrated in Figure \ref{fig:syntheticData}. In the CGP case, the observations are sequentially divided into four segments (each segment has the length of 1024); in the SGP (FITC) case, 128 inducing variables are placed uniformly across the space of observation inputs. 

\begin{figure*}[h]
	\centering
	\subfigure[{$\mathbb{E}[(\mu_{\text{GP}} - \mu_{\text{CGP}})^2]$, $K = 4$}]{\includegraphics[width = .225\linewidth]{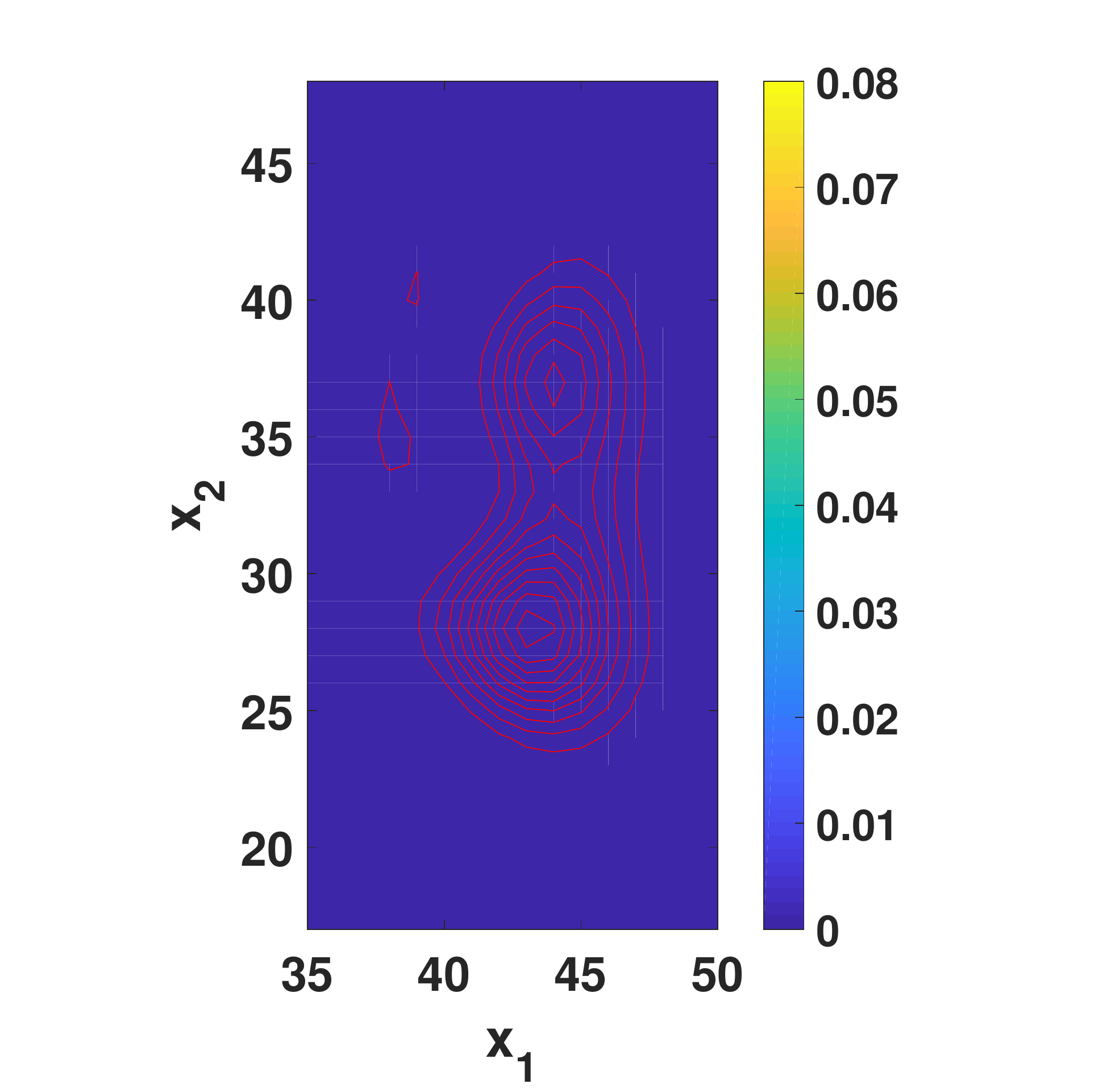}}\qquad
	\subfigure[{$\mathbb{E}[(\mu_{\text{GP}} - \mu_{\text{CGP}})^2]$, $K = 16$}]{\includegraphics[width = .225\linewidth]{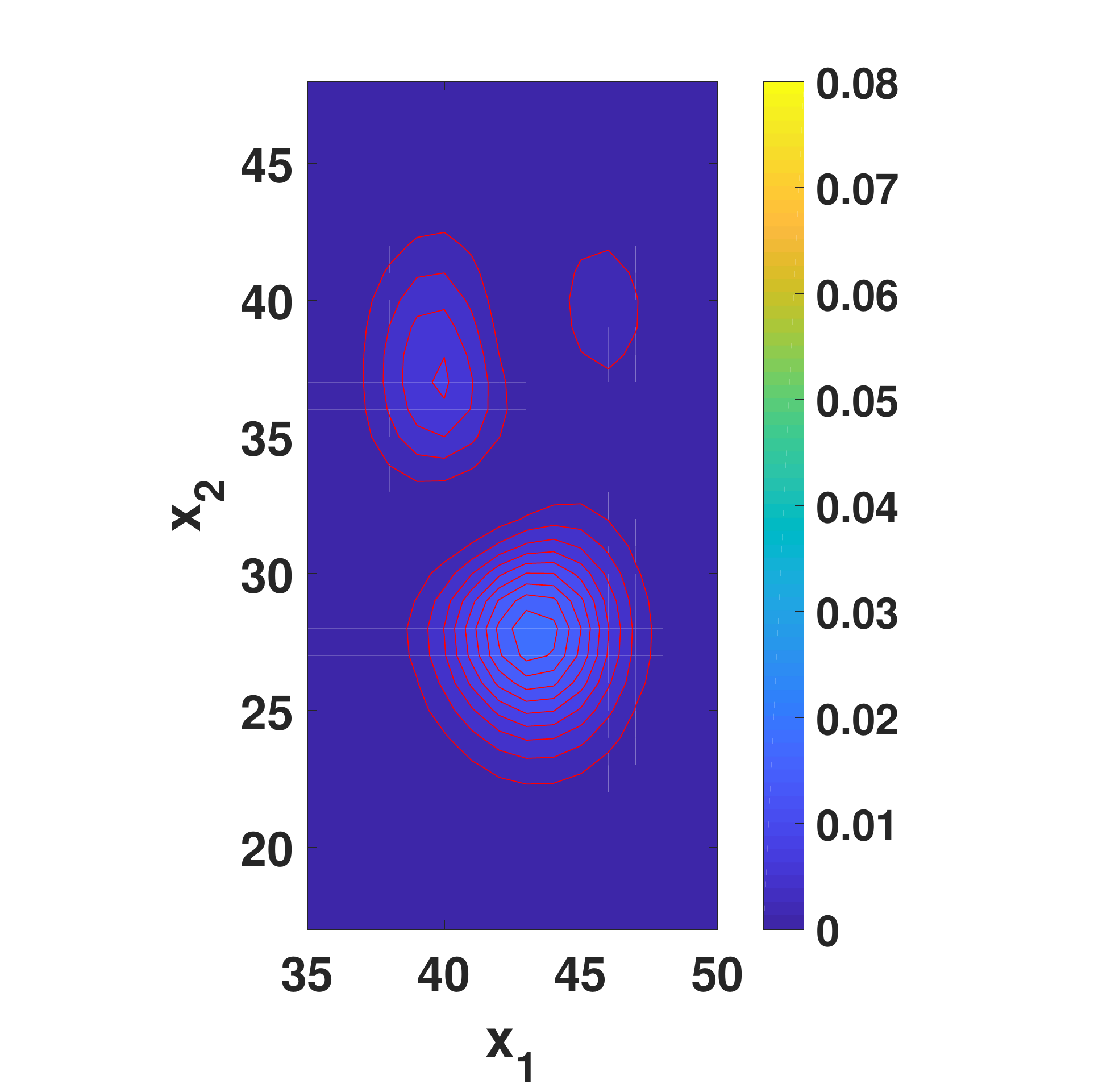}}\qquad
	\subfigure[{$\mathbb{E}[(\mu_{\text{GP}} - \mu_{\text{SGP}})^2]$}]{\includegraphics[width = .225\linewidth]{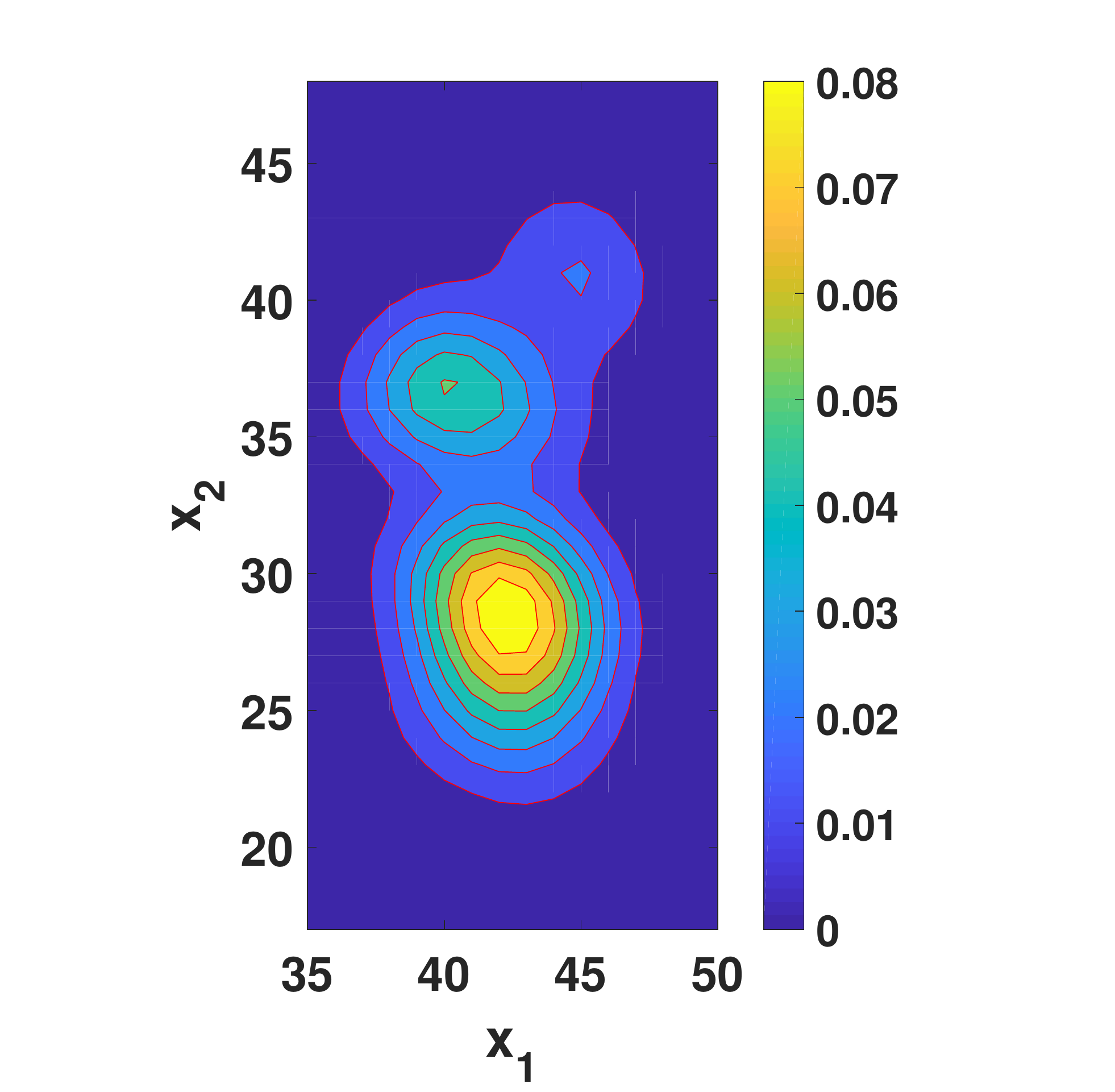}}\qquad
	\caption{Data-averaged approximation errors for CGP ($K = 4$ and $16$) and SGP comparing to GP spatial interpolations.}
	\label{fig:spatial_error}
\end{figure*}

Several observations are made from this example. First, the recursive computations render the CGP very fast comparing to the GP or the SGP methods. Running on a standard laptop, runtime of CGP i less than a second to process 5000 data points and the run time grows linearly with respect to $K$. Second, in this particular case, the predictive variances are underestimated with the composite GP. This is consistent with redundant information in the data segments as per Remark \ref{remark:overAndUnderestimate}. 

\subsection{Synthetic Spatial Data}

In this example, we consider a two-dimensional Gaussian random field (GRF) model: $y(x_1, x_2) \sim \mathcal{GP}(0, k(x_1, x_2, x_1', x_2'))$, where $k(x_1, x_2, x_1', x_2') = \alpha^2\exp[\frac{-(x_1-x_1')^2}{\theta_1^2} + \frac{-(x_2-x_2')^2}{\theta_2^2}]$. This GRF model can be efficiently simulated via circulant embedding \cite{2015-Kroese-SpatialProcessSimulation}. The contour plot of a realization of the GRF ($\alpha = 1, \theta_1 = 8, \theta_2 = 8$) is illustrated in Figure \ref{fig:grf}. This $64\times64$ grid of GRF points are partitioned into $N=3840$ points for prediction as observations and $512$ test points.     

The predictions are illustrated in Figure \ref{fig:spatial}. Visually, CGP is most similar to GP, and the posterior variances of the CGP representing the uncertainty is slightly higher than that of GP. By contrast, SGP and the GP exhibit notable differences and the posterior variance of the former is significantly higher than the latter. The additional MSE \eqref{eq:mse} incurred by using these posterior distributions is visualized for CGP ($K = 4$ and $K = 16$) and SGP in Figure \ref{fig:spatial_error}, using 100 different realizations of the GRF. When the number of data segments varies from 4 to 16, the computation time of CGP is reduced but the additional MSE increases, which is consistent with our analysis in the previous section. By contrast, SGP yields a notably higher additional MSE over GP.

\subsection{A Real-world Case: NOx Prediction}
\begin{figure}[h]
	\centering
	\subfigure[NOx measurements \label{fig:nox}]{\includegraphics[width = .45\linewidth]{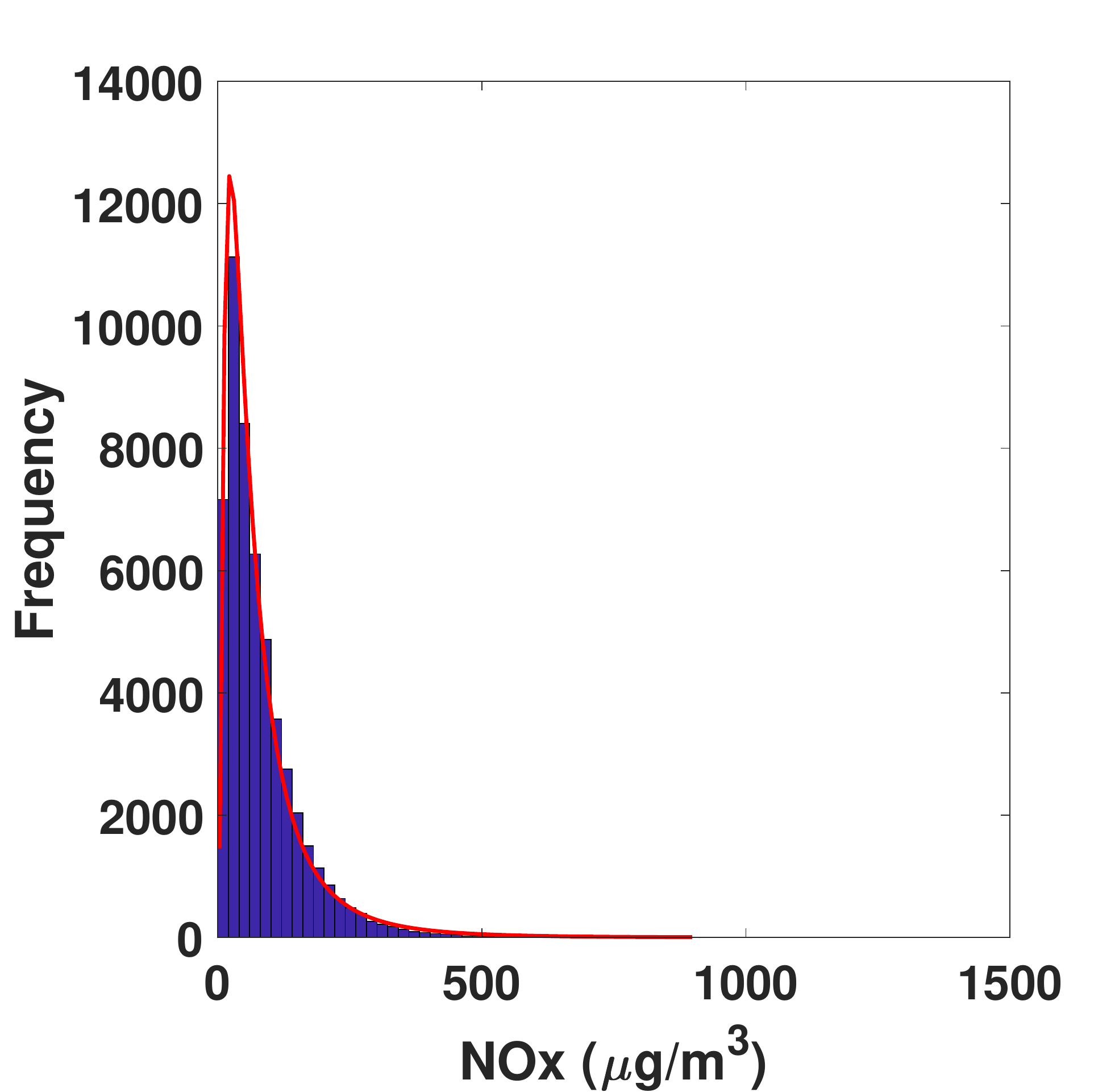}}\qquad
	\subfigure[After log transformation \label{fig:lognox}]{\includegraphics[width = .45\linewidth]{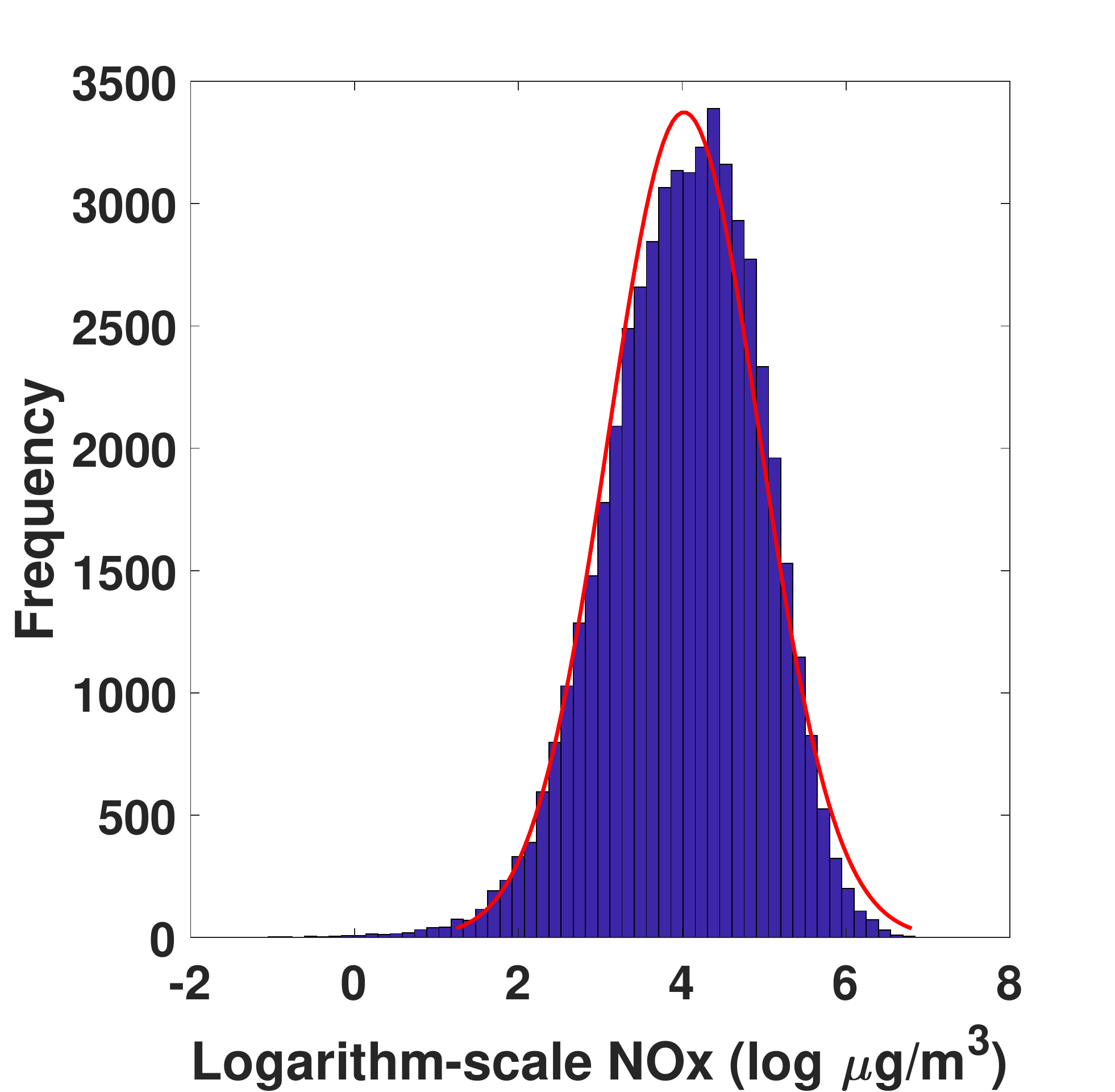}}\qquad
	\caption{Logarithm transformation of NOx measurements.}
\end{figure}

Next we demonstrate the use of CGP in a real application for predicting NOx (nitrogen oxides). The data set (available online: http://slb.nu/slbanalys/) includes more than 10 years hourly NOx measurements for a city with 1 million population. Figure \ref{fig:nox} shows that the NOx data are far from Gaussian: the measurements are non-negative, highly skewed towards to lower values, and have a long tail in high values. Therefore, we apply the logarithm transformation \cite{2011-Shumway-TimeSeries} to adjust the NOx measurements to a normal distributed data. The result after transformation is shown in Figure \ref{fig:lognox}. 

\begin{figure}[h]
	\centering
	\includegraphics[width = .95\linewidth]{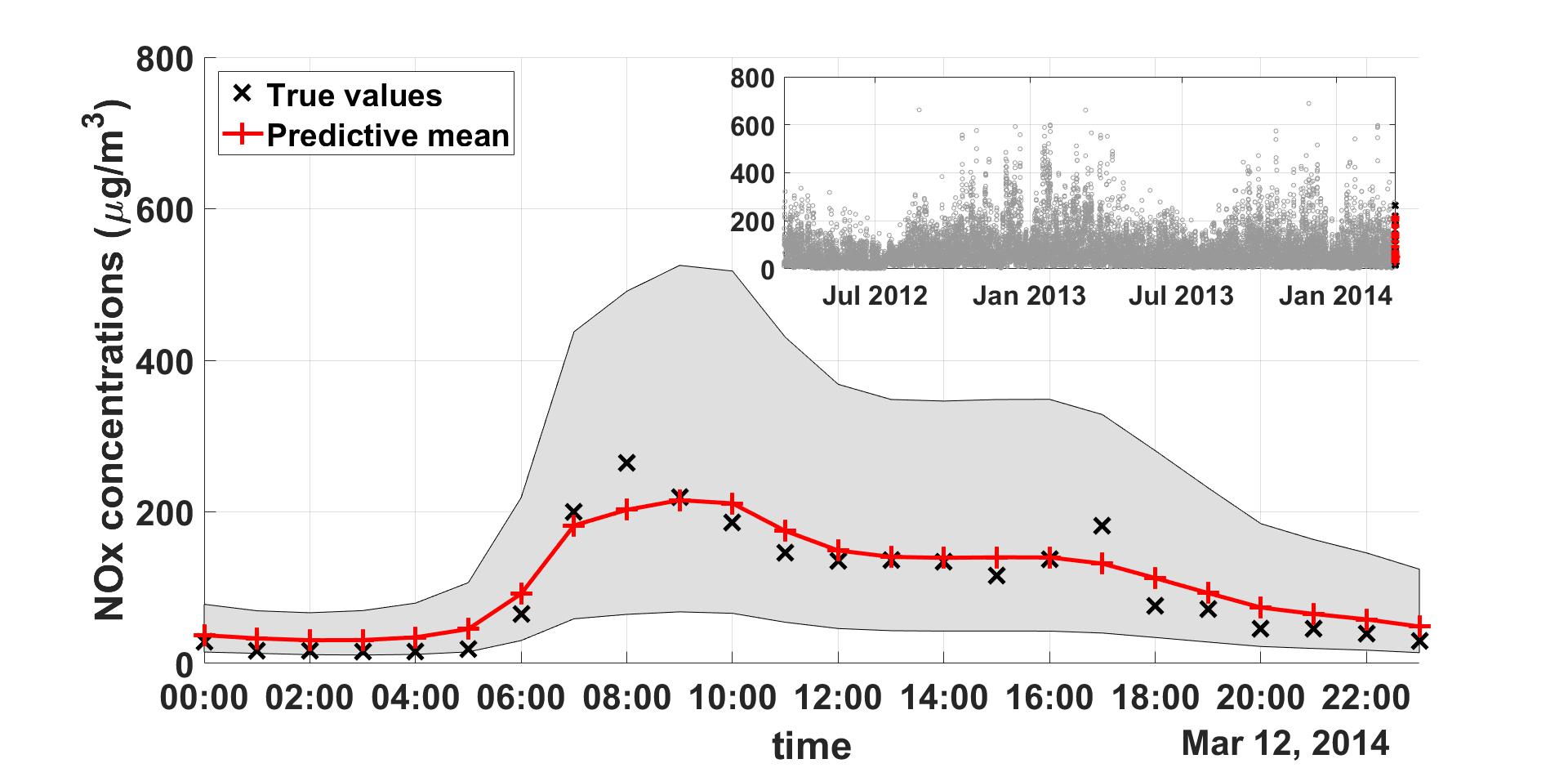}
	\caption{A 24-hour NOx prediction with the composite GP.}
	\label{fig:noxPrediction}
\end{figure} 

An example of 24-hour NOx prediction is illustrated in Figure \ref{fig:noxPrediction}. In this example, previous two years measurements (17472 hourly measurements) are used by the CGP model to produce the posterior of NOx levels on March 12, 2014. The produced posterior is a log-normal process: the predictions are non-negative; the distribution is skewed to the lower values and has a long tail in higher values; and the width of credibility intervals is significantly bigger for rush hours comparing to late and early hours in working days.

\begin{figure}[h]
	\centering
	\includegraphics[width = .95\linewidth]{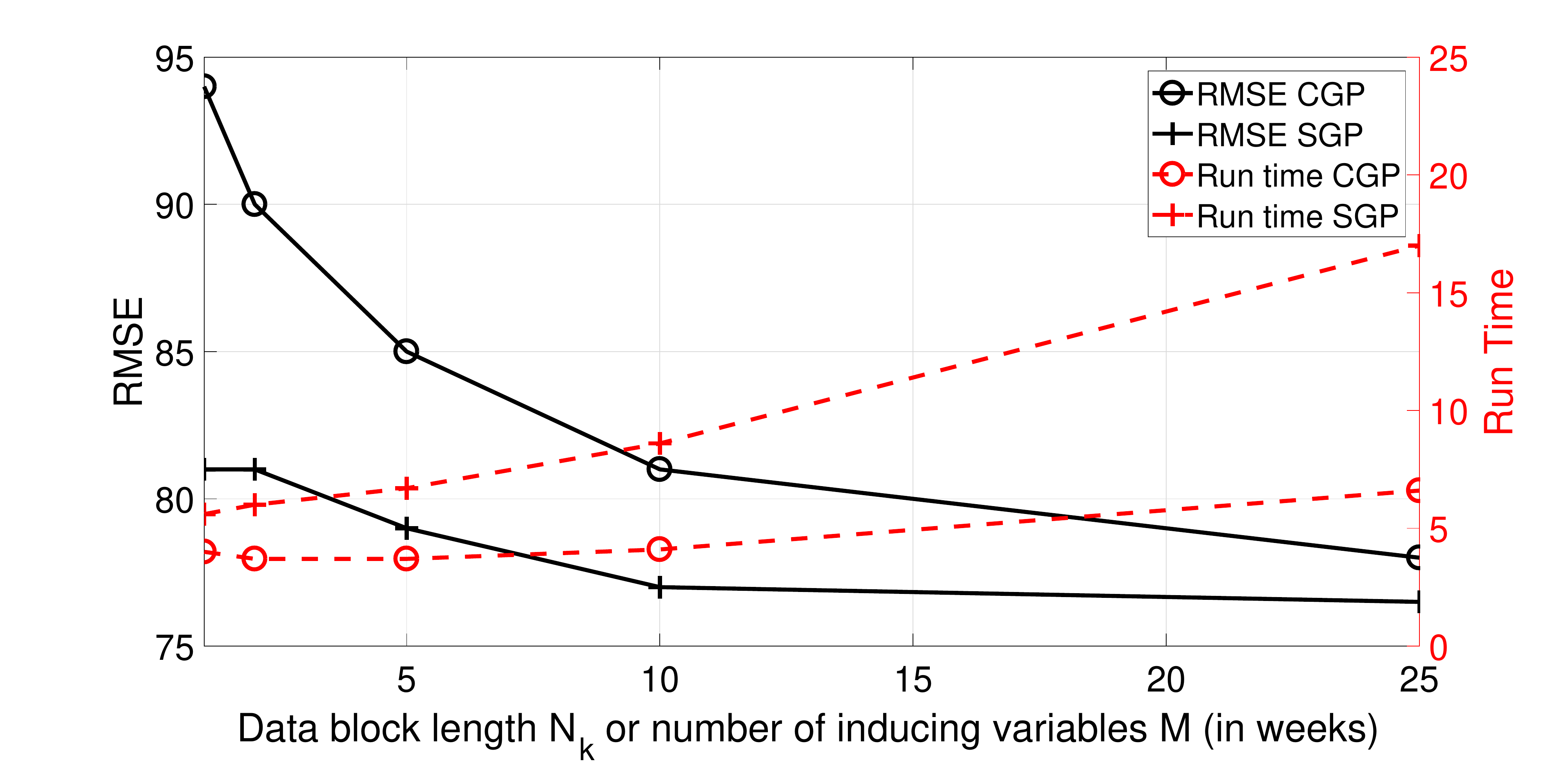}
	\caption{Experiments of daily NOx prediction over year 2014.}
	\label{fig:experiment}
\end{figure}

Finally, we show the root-mean square error (RMSE) and runtime results for the CGP and the SGP, with 365 24-hour predictions over the year of 2014. The GP prediction is used as a reference for comparison. As GP prediction is not very scalable, we limit the maximum number of weeks used for one prediction (for instance the prediction shown in Figure \ref{fig:noxPrediction}) to be 50. Thereafter, we vary the number of observations per batch for the CGP and the number of inducing variables for the SGP from 1 to 25 weeks. The resulting RMSE ($\mu g/m^3$) and runtime (seconds) are shown in Figure \ref{fig:experiment}. The RMSE of GP with 50 weeks of data is slightly above 75, which is provided as the lower bound when comparing the CGP and the SGP results. Figure \ref{fig:experiment} shows a trade-off between the runtime and the RMSE with CGPs and SGPs for this particular data set when the total data used for one prediction is fixed (50 weeks). The SGP with 5 weeks of inducing variable achieves lower RMSE than the CGP with 10 batches and 5 weeks per batch. However, the runtime of the CGP is significantly lower than the SGP. We must notice that when the number of observations per batch or number of inducing variables increases, the difference of runtimes becomes larger and the gap between RMSEs are closing, which shows a nice scalablility and accuracy of the CGP model.


\section{Conclusion}\label{sec:conclusion}
In this work, we addressed the problem of learning GP models and predicting an underlying process when in scenarios where the data sets are large. Using a general belief update framework, we applied a composite likelihood approach and derived the CGP posterior distribution. It can be both be updated and learned recursively with a runtime that is linear in the number of data segments. We show that GP, CGP, and SGP posteriors can be all recovered in this framework using different likelihood models. 

Furthermore, we compared the CGP posterior with that of GP. We obtained closed-form expressions of the additional prediction MSE incurred by CGP and the differences in information gains under both models. The results can be used as a conceptual as well as computational tool for designing segmentation schemes and evaluating errors induced when using the scalable CGP in different applications. The design of segmentation schemes based on the derived quantities is a topic of future research.

\clearpage
\bibliography{references}
\bibliographystyle{icml2018}

\end{document}